\pdfoutput=1

\documentclass[11pt]{article}

\usepackage[]{acl}

\usepackage{times}
\usepackage{latexsym}

\usepackage[T1]{fontenc}

\usepackage[utf8]{inputenc}

\usepackage{microtype}

\usepackage{enumitem}
\setitemize{noitemsep,topsep=0pt,parsep=0pt,partopsep=0pt}
\definecolor{cite}{rgb}{0.6,0.6,1.0}

\definecolor{todo}{rgb}{1,0.5,0}

\definecolor{review}{rgb}{0,0.7,0}

\usepackage{booktabs}
\usepackage{graphicx}
\usepackage{float}
\usepackage{tabularx}
\usepackage{soul}

\usepackage{multirow}
\newcommand{\mr}[2]{\multirow{#1}{*}{#2}} 
\newcolumntype{Y}{>{\centering\arraybackslash}X} 

\usepackage{xcolor}
\usepackage{mdframed}

\usepackage{threeparttable} 

\usepackage{caption}
\usepackage{subcaption}
\usepackage{arydshln}
\usepackage{xspace}
\usepackage{csquotes}
\usepackage{pifont}
\usepackage{makecell}

\newcommand{\checkmark}{\ding{51}}%
\newcommand{\xmark}{\ding{55}}

\newcommand{\XLMRBase}{\mbox{XLM-R\textsuperscript{Base}}\xspace}
\newcommand{\XLMRDomain}{\mbox{XLM-R\textsuperscript{Domain}}\xspace}
\newcommand{\MADXDomain}{{MAD-X+Domain}\xspace}
\newcommand{\MADXSquare}{{MAD-X\textsuperscript{2}}\xspace}

\definecolor{darkgreen}{RGB}{5,180,0}

\title{M2QA: Multi-domain Multilingual Question Answering}

\author{
\bf Leon Engl\"ander\thanks{\, Authors contributed equally.}\hspace{1.3mm}$^1$, Hannah Sterz$^{*2}$, Clifton Poth$^{4}$,  \\ 
{\bf Jonas Pfeiffer\thanks{\,\,Jonas did not run any experiments or use the data in this paper. No other GDM/Google employee was part of this project. No GDM resources were used in this project.}\hspace{1.3mm}$^{3}$
, Ilia Kuznetsov$^{1}$, Iryna Gurevych$^{1}$} \\
$^1$Ubiquitous Knowledge Processing Lab,  Technical University of Darmstadt \\  
$^2$Language Technology Lab, University of Cambridge \hspace{0.5em} \\
$^3$Google DeepMind \hspace{0.5em} $^4$Cohere\\
\url{www.ukp.tu-darmstadt.de}}

\begin{document}
\maketitle
\begin{abstract}
Generalization and robustness to input variation are core desiderata of machine learning research. Language varies along several axes, most importantly, language instance (e.g. French) and domain (e.g. news). While adapting NLP models to new languages within a single domain, or to new domains within a single language, is widely studied, research in joint adaptation is hampered by the lack of evaluation datasets. This prevents the transfer of NLP systems from well-resourced languages and domains to non-dominant language-domain combinations. 
To address this gap, we introduce M2QA, a multi-domain multilingual question answering benchmark.
M2QA includes 13,500 SQuAD 2.0-style question-answer instances in German, Turkish, and Chinese for the domains of product reviews, news, and creative writing. We use M2QA to explore cross-lingual cross-domain performance of fine-tuned models and state-of-the-art LLMs and investigate modular approaches to domain and language adaptation.
We witness  \textbf{1)} considerable performance \textit{variations} across domain-language combinations within model classes and \textbf{2)} considerable performance \textit{drops} between source and target language-domain combinations across all model sizes. We demonstrate that M2QA is far from solved, and new methods to effectively transfer both linguistic and domain-specific information are necessary.\footnote{We make M2QA publicly available at \url{https://github.com/UKPLab/m2qa}}

\end{abstract}
\begin{figure}
    \centering
    \includegraphics[width=0.8\linewidth]{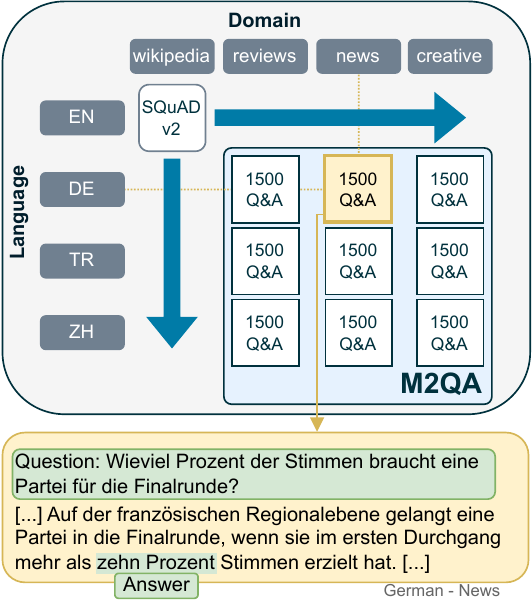}
    \vspace{-1.5mm}
    \caption{M2QA enables joint multi-domain and multilingual QA evaluation of NLP models across three diverse languages and three distinct domains (top) with 1500 SQuAD 2.0-style question-answer pairs for each language-domain combination (bottom).}
    \label{fig:eye_catcher}
    \vspace{-1em}
\end{figure}

\section{Introduction}
One of the central goals of natural language processing (NLP) is to develop systems that generalize well across different distributions, such as texts in different languages and domains.\footnote{Domains defined as text associated with a specific topic, such as product reviews or news \cite{gururangan-etal-2020-dont}.} While Transformer models have brought tremendous progress in NLP in recent years, especially evident with the recent emergence of large language models (LLMs), the problem of generalizing to new domains and languages is still far from solved. In-context learning (ICL), which refers to the ability of LLMs to perform tasks based on examples or instructions in the input prompt \cite{Brown2020Language}, is likely the reason for their emergent abilities \cite{lu2023emergent} -- yet, even with in-context learning, Transformers cannot generalize beyond their pre-training data \cite{yadlowsky2023pretraining}, and their performance varies considerably across languages and is particularly low in languages underrepresented in the training data \cite{laskar-etal-2023-systematic}.

With over 7,000 documented languages \cite{joshi-etal-2020-state} and countless domains, ensuring sufficient pre-training data coverage for every possible language-domain pair is hardly feasible.
This motivates the development of methods that allow NLP systems to adapt to new languages and domains.
While isolated language \cite[e.g.][]{conneau-etal-2018-xnli, hu2020xtreme, artetxe-etal-2020-cross, ghaddar-langlais-2017-winer, scialom-etal-2020-mlsum} and domain adaptation \cite[e.g.][]{wang-etal-2018-glue} are extensively covered in prior work, the lack of comprehensive multi-domain multilingual benchmarks makes it difficult to objectively evaluate joint language and domain transfer methods. Existing multi-domain multilingual benchmarks either contain only one language in addition to English \cite{gupta-etal-2018-mmqa}, use machine-generated text \cite{bassignana-etal-2023-multi} or are task-oriented dialogue systems that use narrow application-specific domains rather than a diverse set of domains useful for a wide range of applications \cite{moghe-etal-2023-multi3nlu, hu-etal-2023-multi-3}. 
Results on these benchmarks suggest that language and domain are not independent axes. Therefore, we cannot infer the performance of joint transfer from individual axes, making it hard to systematically compare NLP models across languages and domains and to study joint language and domain adaptation approaches.

To address this gap, we introduce M2QA, a multi-domain multilingual SQuAD 2.0-style \cite{rajpurkar-etal-2016-squad} extractive question answering (QA) dataset.
We manually annotate naturally occurring texts in the respective languages -- as opposed to translating documents from English -- in order to increase lexical diversity \cite{rabinovich-etal-2016-similarities}, mitigate the introduction of artifacts \cite{artetxe-etal-2020-translation} such as  ``translationese'' \cite{bizzoni-etal-2020-human}, and integrate the cultural idiosyncrasies of the target language \cite{hershcovich-etal-2022-challenges,kuulmets-fishel-2023-translated}.
The new benchmark makes it possible to study how well existing models perform at joint language and domain transfer (RQ1), whether specific language-domain combinations are especially hard to tackle for current models (RQ2), and whether existing methods (e.g. full fine-tuning,  modular setups, ICL / instruction-based methods for LLMs)  compare on language and domain transfer (RQ3). 

In summary, our paper makes the following contributions:
\textbf{1)}~We create a multi-domain multilingual extractive QA benchmark, covering three domains (product reviews, news, creative writing) and three languages (German, Turkish, Chinese), resulting in 13,500 answerable and unanswerable QA instances (Figure \ref{fig:eye_catcher}).
\textbf{2)}~We evaluate baseline and transfer performance using a wide range of models and transfer techniques, including fully-finetuned models, modular transfer learning and LLMs.
\textbf{3)}~We find that transfer performance considerably varies across domain-language combinations.
\textbf{4)}~We find that the widely used SQuAD 2.0 evaluation metric implementation is insufficient for evaluating multilingual extractive QA due to its reliance upon whitespace tokenization and propose a version of the metric that mitigates the issue.
\textbf{5)}~Our results show that modern LLMs perform considerably worse on their target than on their source domain-language pair, highlighting the need for further research into methods that transfer both linguistic \emph{and} domain-specific information.

\section{Background}
\subsection{Adaptation and Modularity}
Transfer and adaptation methods aim to optimize model performance on unseen data distributions. This can be achieved through modular deep learning methods \cite{pfeiffer2023modular} that combine modules containing knowledge about different aspects of the task, such as the language or domain.
Modular approaches may involve merging weights of individually trained models \cite{ilharco2022editing} or model ensembling \cite{blevins2024breaking, li2022branch}. However, these methods require fully fine-tuning multiple models.
Parameter-efficient fine-tuning (PEFT) or adapter\footnote{We use the terms ``adapter'' and ``PEFT'' interchangeably.} methods \cite{houlsby2019parameter, hu2021lora, ben-zaken-etal-2022-bitfit, ansell-etal-2022-composable} overcome these limitations. Instead of updating all model weights in the fine-tuning stage, adapter methods only fine-tune a small set of parameters while keeping the majority of parameters frozen.

\subsubsection{Domain and Language Transfer}
Domain transfer is the process of learning a task on a set of domains and then applying the model to the same task in a previously unseen domain. Domain transfer can be accomplished by (sequentially) fine-tuning LMs on in-domain data \citep{howard-ruder-2018-universal,pruksachatkun-etal-2020-intermediate,poth-etal-2021-pre, gururangan-etal-2020-dont} or by combining multiple expert LMs \cite{li2022branch}, where new domains are added by training new expert LMs. \citet{gururangan2023scaling} propose to cluster the data in the beginning to avoid massive node synchronization. 
To avoid full model fine-tuning, domain-specific adapters have been used \cite{chronopoulou-etal-2022-efficient, chronopoulou-etal-2023-adaptersoup}. \citet{gururangan-etal-2022-demix} replace the Transformer's feedforward layers with DEMIX layers consisting of multiple domain experts. 
In this modular solution, the DEMIX layers of different domains can be combined to handle heterogeneous domains during inference.

For language transfer, a model trained on a task in one or more source language(s) is evaluated on a different target language. While LLMs such as GPT-4 \cite{openai2023gpt4} can perform zero-shot or few-shot language transfer between similar languages, smaller models -- such as XLM-Roberta \citep{conneau-etal-2020-unsupervised} -- need to be fine-tuned or otherwise adapted: \citet{blevins2024breaking} combine multiple expert LMs, similar to the domain branch-train-merge setup; modular approaches \cite[e.g.][]{pfeiffer-etal-2020-mad, ansell-etal-2021-mad-g, parovic-etal-2022-bad, parovic-etal-2023-cross} train language-specific and task-specific adapters to perform language transfer by exchanging the language adapter. 
Modular setups also find application in the joint transfer between domain and language. \citet{cooper-stickland-etal-2021-multilingual} use domain and language-specific adapters to transfer to languages and domains. The m$^4$ adapter \cite{lai-etal-2022-4} uses meta-learning with adapters for multi-domain multilingual machine translation. \citet{kulkarni-etal-2023-towards} propose a mixture-of-experts to perform multi-domain multilingual named entity recognition. 

In this paper, we explore variations of all previously mentioned adaptation techniques: \textbf{1)}~fully fine-tuning smaller models; \textbf{2)}~a modular setup following the MAD-X method;~\textbf{3)} zero-shot and few-shot approaches using LLMs.

\subsection{Evaluation}
Domain and language transfer techniques are mainly evaluated based on perplexity \cite[e.g.][]{li2022branch, gururangan-etal-2022-demix, conneau2019cross} or downstream tasks \cite[e.g.][]{pfeiffer-etal-2020-mad, gupta-etal-2023-cross}. Perplexity is a token-level metric which overemphasizes the importance of frequent tokens and constructions \cite{dudy-bedrick-2020-words} and does not necessarily account for task-specific phenomena. Hence, it is questionable if perplexity is a good indicator of downstream task performance. 



Question answering has been used to evaluate cross-lingual or cross-domain transfer separately. Prominent multilingual datasets are XQuAD \cite{artetxe-etal-2020-cross} and MLQA \cite{lewis-etal-2020-mlqa}.
For domain transfer, the Quail benchmark \cite{rogers2020getting} provides a multiple-choice QA dataset. MultiReQA \cite{guo-etal-2021-multireqa} combines existing QA datasets to a new multi-domain benchmark. 


Benchmarks that target cross-lingual and cross-domain transfer in other tasks than QA also exist;
MultiFC \cite{augenstein-etal-2019-multifc} and  CrossRE \cite{bassignana-plank-2022-crossre} contain multiple domains for the same task.
M2D2 \cite{reid-etal-2022-m2d2} introduces a massively multi-domain setup with 145 subdomains evaluating performance with perplexity. \citet{chronopoulou-etal-2022-efficient} evaluate perplexity across domains found on websites. Other popular cross-lingual tasks are NER \cite[e.g.][]{ghaddar-langlais-2017-winer} and summarization \cite[e.g.][]{scialom-etal-2020-mlsum}.
Most NLP benchmarks only focus on exploring one dimension, i.e. multilinguality \textit{or} multi-domain \cite{ruder-etal-2022-square}, which prevents investigating non-linear dependencies between domain and language.
We discuss this in more detail in Section \ref{sec:m2qa_dataset_requirements} below.


\section{M2QA Dataset}
\subsection{Requirements}
\label{sec:m2qa_dataset_requirements}
We define the following requirements for a benchmark that allows joint evaluation of language and domain transfer methods: (R1) Coverage: The benchmark should provide annotated data for each language-domain combination. (R2) Diversity: The benchmark should cover typologically distinct language and a broad range of domains. (R3) Openness: The source texts should be open-licensed and available for research usage. (R4) Universal task: The data should be annotated using a domain-agnostic task, enabling cross-domain comparison.

An additional and important trade-off pertains to the use of translated vs. naturally occurring texts. Translated texts ensure that the data covers the same topics within the domain, resulting in aligned text across the languages. However, translations have lower lexical diversity \cite{rabinovich-etal-2016-similarities} and introduce artifacts \cite{artetxe-etal-2020-translation} such as unnatural language usage and  ``translationese'' \cite{bizzoni-etal-2020-human}. \citet{hershcovich-etal-2022-challenges} show that culture affects several axes of text variation. Translations contain the cultural background of the source language that does not correspond to the cultural background of native speakers of the target language \cite{kuulmets-fishel-2023-translated}.
We prioritize language representative of how native speakers write over aligned text. Thus, we require (R5) Naturalness: all texts in the benchmark should have been produced naturally, not via translation.

Few multilingual and multi-domain datasets have been previously proposed. MMQA \cite{gupta-etal-2018-mmqa} includes factoid and short descriptive questions in English and Hindi over 6 domains. 
Multi3WOZ \cite{Hu2023Multi3WOZ} and Multi3NLU++ \cite{moghe-etal-2023-multi3nlu} are multi-domain and multilingual benchmarks for task-oriented dialogue. 
README++ \cite{naous2023readme} is a multi-domain multilingual benchmark for readability assessment which includes translated texts in some of the domains.
CrossRE \cite{bassignana-plank-2022-crossre} is a machine-generated, human-verified, multi-domain, multilingual benchmark for relation extraction.
As Table \ref{tab:existing_datasets} shows, none of the existing datasets fulfil our requirements as defined above.  

\begin{table*}[]
    \small
    \centering
    \resizebox{0.99\textwidth}{!}{
    \begin{tabular}{lcccccc}
        \toprule
         Dataset & Task & \makecell{Coverage (R1)} &  \makecell{Diversity (R2)}& \makecell{Openness (R3)}  & \makecell{Universal Task (R4)} & \makecell{Naturalness (R5)} \\
         \midrule
         MMQA  \cite{gupta-etal-2018-mmqa}  & QA & \checkmark  & \xmark & \checkmark  & \checkmark  & \checkmark \\
         Multi3WOZ  \cite{Hu2023Multi3WOZ} & ToD & \checkmark & \checkmark  & \checkmark & \xmark & \checkmark \\
         Multi3NLU++ \cite{moghe-etal-2023-multi3nlu} & ToD & \checkmark & \xmark & \checkmark & \xmark &  \xmark\\
         README++  \cite{naous2023readme} & RA & \xmark & \checkmark & \checkmark & \xmark & (\checkmark ) \\
         CrossRE  \cite{bassignana-plank-2022-crossre} &  RE &  \checkmark & \checkmark & \checkmark & \checkmark & \xmark \\
         \midrule
         M2QA & QA & \checkmark  & \checkmark & \checkmark & \checkmark & \checkmark \\ 
         \bottomrule
    \end{tabular}
    }
    \vspace{-1mm}
    \caption{Overview of existing multilingual multi-domain datasets along with their key characteristics and task. (QA = Question Answering, ToD = Task-oriented Dialogue, RA = Readability Assessment, RE = Relation Extraction).}
    \label{tab:existing_datasets}
    \vspace{-2.5mm}
\end{table*}

\subsection{Design}


As per our requirements, the languages and domains in M2QA should cover a variety of language families and text styles (R2) to ensure that the transfer is not trivial.
We chose German (Indo-European Germanic), Turkish (Turkic), and Chinese (Sino-Tibetan) as languages.
As domains, we chose product reviews, news, and creative writing, covering various writing styles, levels of formality, and vocabularies. 
To fulfil R1, we annotated data for every language and domain combination. We collected open (R3) texts that are originally written in the target language to ensure naturalness (R5). 

The annotated task needs to be universal (R4). 
One universal task is extractive question answering (QA). For extractive QA, the input is a question and a context that provides information to answer the question. The task is to extract the shortest span from the context that answers the question or, if the context does not contain an answer to the question, return that the question is unanswerable. 
An example question is shown in Figure \ref{fig:eye_catcher}. Extractive QA requires natural language understanding to identify the information needed to answer the question. Additionally, it requires reasoning to connect the concepts mentioned in the question to those mentioned in the text and extract the span with the relevant information. This makes extractive QA a complex task suitable for our benchmark.

\subsection{Dataset Creation}
Our annotation process consists of three parts: Passage curation, annotation and quality assurance.

\paragraph{Passage Curation.}
Collecting a benchmark that contains multiple languages and domains is not trivial, as the language and domain are entangled. Additionally, the data size varies for different domain and language combinations. For instance, scientific texts are mostly written in English. During the creation of the M2QA benchmark, we collected task annotations from combinations that are non-trivial to find. For instance, with our requirement for the data to be open (R3), finding creative writing data is challenging as most books have a copyright.
For product reviews, we use the Chinese and German parts of MARC \cite{keung-etal-2020-multilingual} and the Turkish product reviews dataset.\footnote{\url{https://huggingface.co/datasets/turkish_product_reviews}}
For news, we use the German 10kGNAD \cite{schabus2017one} dataset, the Chinese CNewSum \cite{Wang2021CNewSum}, and Turkish BilCat \cite{Toraman2011Developing}.
The creative writing domain is covered by German books from the Gutenberg Corpus \cite{Gerlach2018ASP} and Turkish and Chinese stories published on Wattpad\footnote{\url{https://www.wattpad.com/}} with an open license.
For more details on the data sources, licensing information, and preprocessing, see Appendix \ref{sec:appendix-passage-curation}.

\paragraph{Annotation.}
For the question-answer collection, we hired crowdworkers from Prolific\footnote{\url{https://www.prolific.com}}, which was chosen due to its high annotation quality \cite{douglas_data_2023} and advanced annotator filtering options. For each passage, the crowdworkers provided three answerable and two unanswerable questions. For answerable questions, they selected the shortest text span of the passage that answers the question.
Following SQuAD 2.0 \cite{rajpurkar-etal-2018-know}, we also let crowdworkers select a plausible answer span for unanswerable questions to make them harder to classify.
We limit the maximum answer length to fall within 97\% of the answers in XQuAD: ten words for German, nine for Turkish, and 22 characters for Chinese. 
For details on the annotation process, see Appendix \ref{sec:appendix-platform-ui}.

\paragraph{Quality Assurance.}
To promote high data quality, crowdworkers were required to have at least a Bachelor's degree, speak the language in which they annotate data as their first language, and be fluent in English to understand the tutorial.\footnote{The tutorial can be found here: \url{https://github.com/UKPLab/m2qa/tree/main/Website}}
After the first annotation session, we manually reviewed ten randomly sampled question-answer pairs for each annotator, including at least one answerable and one unanswerable question. We translated annotations with DeepL.\footnote{\url{https://www.deepl.com/api}}
If more than one QA pair violated our guidelines, we excluded the annotator's data from the dataset and removed the annotator from the worker pool. The results of the manual checks can be found in Appendix \ref{sec:appendix:annotation_process_quality}.
In total, we employed 162 crowdworker annotators, of which $19\%$ (31 annotators) were rejected for poor-quality questions. 
From the questions kept, we manually checked 1310 questions ($9.7\%$ of the dataset)

\subsection{Statistics}
We collected 1500 question-answer pairs for every domain-language combination, resulting in 13,500 question-answer pairs.
The domains are lexically diverse: maximum Jaccard similarity between domains is 0.135 in German, 0.115 in Turkish and 0.169 in Chinese (Appendix \ref{sec:appendix:passages_lexical_diversity}). The average answer length is 3.62 words in German, 3.06 in Turkish and 4.46 in Chinese, similar to XQuAD \cite{artetxe-etal-2020-cross} with 2.98 words in German, 2.92 in Turkish, and 3.51 in Chinese respectively.


\section{Experiments}
The curation of the M2QA benchmark allows us -- for the first time -- to explore the transfer capabilities of state-of-the-art LMs along multiple dimensions.
We will use M2QA to investigate the following research questions: \textbf{(RQ1)} How well do existing models perform at transfer learning across language and domains jointly?
\textbf{(RQ2)} What language domain combinations are especially hard to tackle for the current models?
\textbf{(RQ3)} How do modular adapter-based methods compare to fully-finetuned models in domain and language transfer?

\subsection{Base Models}
We first introduce our baseline models. See Appendix \ref{sec:appendix-baseline-training} for details on \mbox{XLM-R models}; Appendix \ref{sec:appendix-prompt-details} lists the LLM prompts.

\vspace{-1mm}
\paragraph{\XLMRBase}
\cite{conneau-etal-2020-unsupervised} is a multilingual Transformer encoder based on RoBERTa \cite{Liu2019RoBERTaAR} that has been extensively studied in prior research on adaptation. We fine-tune the model on the English Wikipedia SQuAD 2.0 dataset \cite{rajpurkar-etal-2018-know} and evaluate it on different languages and domains of the M2QA benchmark. For data samples from languages other than English and not from the Wikipedia domain, this requires transfer across both dimensions. 

\vspace{-1mm}
\paragraph{\XLMRDomain}
As a second baseline, we evaluate XLM-R in a cross-lingual but not cross-domain transfer setup. For each domain, we first train an individual XLM-R model on domain-specific texts in English (see Appendix \ref{table:appendix:english-domain-datasources}) for 100,000 update steps via Masked Language Modeling (MLM). After this intermediate domain fine-tuning, we fine-tune the domain-adapted models on the SQuAD 2.0 dataset.

\vspace{-1mm}
\paragraph{LLaMA}
We evaluate the performance of Llama 2-chat 13B \cite{touvron-etal-2023-llama}\footnote{Mostly trained on English ($89.7 \%$ of the training data).} and Llama 3-instruct 8B \cite{llama3modelcard}. We apply simple postprocessing to extract the answer from the generated text; see Appendix \ref{sec:appendix:llm-postprocessing} for details.

\vspace{-1mm}
\paragraph{GPT-3.5}
We also experiment with GPT-3.5 \cite{Brown2020Language}. As its behavior changes over time \cite{chen2023chatgpts}, we investigate two versions of \texttt{gpt-3.5-turbo}: \texttt{-0301}  and \texttt{-0613} 

\vspace{-1mm}
\paragraph{Aya 23}
Lastly, we evaluate Aya 23 8B \cite{aryabumi2024aya}, a multilingual large language model.

\subsection{Setup}
\begin{figure}[t]
    \centering
    \includegraphics[width=\linewidth]{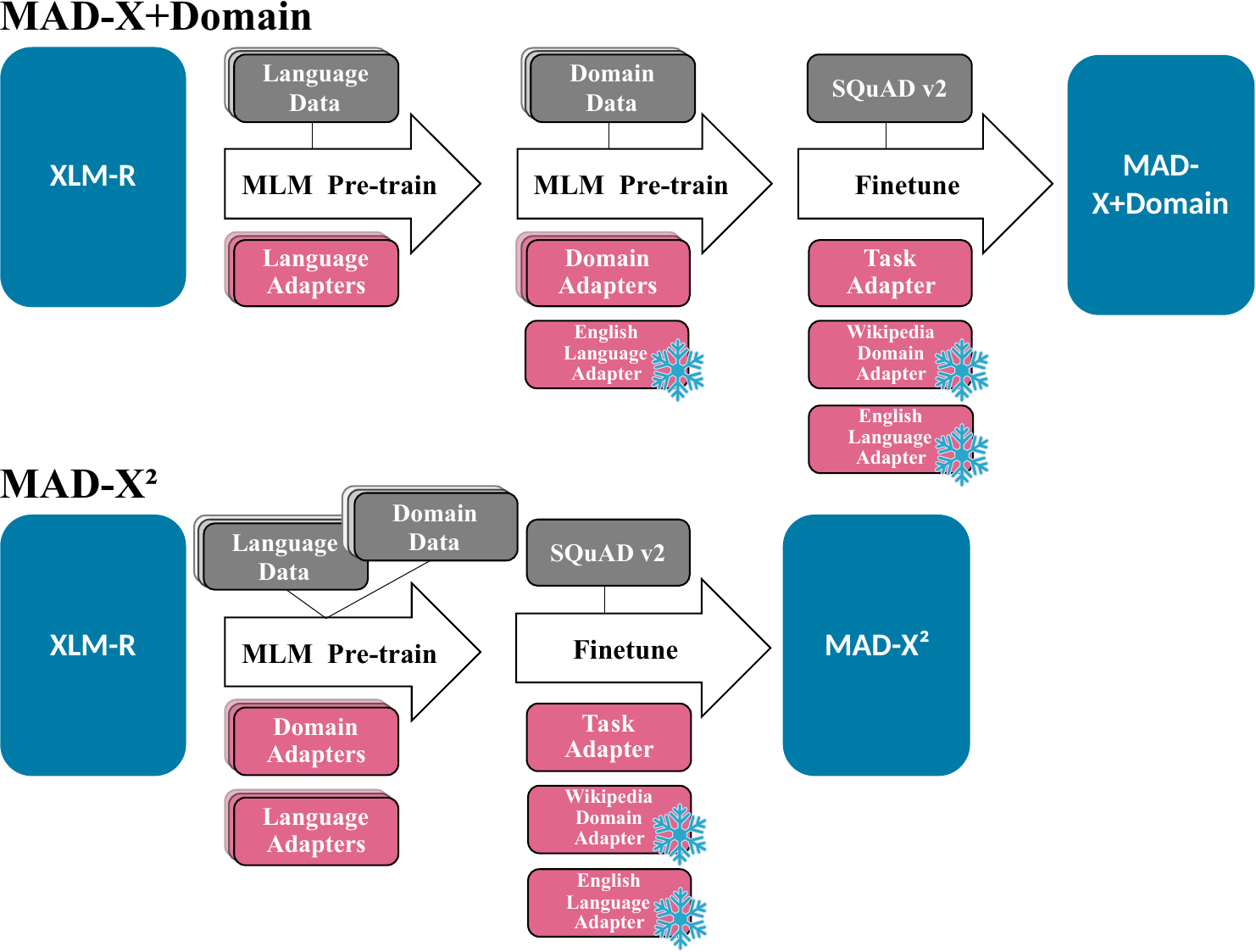}
    \caption{The training process for the modular setups.}
    \label{fig:madXsetups}
\end{figure}

Here we introduce a new modular setup that extends MAD-X for language and domain transfer. We propose two training variants: \MADXDomain and \MADXSquare. Figure \ref{fig:madXsetups} illustrates the approaches, and Appendix \ref{sec:appendix-mad-xexperiments-traininig} provides details.

\vspace{-1mm}
\paragraph{\MADXDomain} We extend the MAD-X \cite{pfeiffer-etal-2020-mad} language transfer framework with a domain adapter by stacking the task adapter above the domain adapter, which is stacked above the language adapter.
We train new domain adapters and use MAD-X's language adapters that were trained via MLM on Wikipedia.
Each domain adapter is trained for 100,000 update steps on the same English domain texts as \XLMRBase using MLM with an activated English language adapter. Then, we train the QA task adapter on SQuAD 2.0 with the English language and Wikipedia domain adapter enabled. During evaluation, we activate the domain and language adapters of the target task.


\vspace{-1mm}
\paragraph{\MADXSquare} The \MADXSquare setup maintains the MAD-X+Domain's adapter architecture but alters the training approach to \emph{simultaneously} train language and domain adapters. We use MLM on texts for every domain-language combination except for Chinese and Turkish creative writing where massive, open-licensed data is scarce.
\footnote{The Chinese and Turkish creative writing test sets in M2QA were manually curated to ensure quality and exclude harmful content -- yet sanitizing a large-scale corpus in such way was not feasible.} During training, we change the domain and language to be trained in each batch, i.e., each batch has text from a different domain and language and the corresponding adapters are activated. We hypothesize that this fosters distinct encapsulation of language-specific and domain-specific knowledge within the respective adapters. We train every domain and language adapter for 62,500 update steps with a batch size of 16, resulting in a total of 250,000 update steps.

\subsection{Results}
\label{sec:results}

We report the performance by language for the answerable and unanswerable questions of M2QA in Table \ref{tab:full_results}, using the F1 and Exact Match (EM) scores as defined by \citet{rajpurkar-etal-2018-know}. For answerable questions, we report both scores, whereas for unanswerable questions, only the F1 score is included since both scores are identical by definition.

\begin{table*}[t]
    \centering
    \resizebox{\textwidth}{!}{%
    \begin{tabular}{llcccccccccccccc}
        \toprule
         &   & \multicolumn{3}{c}{\textbf{Creative Writing}} & \multicolumn{3}{c}{\textbf{Product Reviews}}  & \multicolumn{3}{c}{\textbf{News}} & \multicolumn{5}{c}{\textbf{Average}}   \\
                    & & \multicolumn{2}{c}{answerable} & unansrbl& \multicolumn{2}{c}{answerable} & unansrbl& \multicolumn{2}{c}{answerable} & unansrbl& \multicolumn{2}{c}{answerable} & unansrbl & \multicolumn{2}{c}{Total Average}\\
         & Model      & F1    & EM & F1   & F1    & EM & F1    & F1    & EM & F1    & F1    & EM & F1    & F1 & EM\\
         \midrule
         \multirow{9}{*}{\rotatebox[origin=l]{90}{\textbf{German}}} & \XLMRBase  & 30.42& 17.67&                      67.83& 35.64& 21.22&                      56.67& 40.98& 26.56&          55.33& 35.68& 21.82&    59.94&  45.38&         37.07\\
         & \XLMRDomain & 18.39& 9.44&                      79.00& 30.41& 17.00&                      60.83& 20.79& 11.56&          69.50& 23.20& 12.67&    69.78&   41.83&        35.51\\
         \cmidrule(lr){2-2}\cmidrule(lr){3-5} \cmidrule(lr){6-8} \cmidrule(lr){9-11} \cmidrule(lr){12-16}
         & MAD-X+Domain & 4.25&2.44& \textbf{94.83} &23.44&13.56&                          73.33&38.82&24.44&              55.33&22.17&13.48&  74.50&  43.19&   37.98\\
         & MAD-X$^2$  &19.09&11.33&   82.33& 22.96&13.44&      72.33&42.59&27.67&53.50&28.21&17.48&     69.39&   44.68&   38.24\\
         \cmidrule(lr){2-16}
         & Llama 2-chat (13b) & 31.19 & 11.89 & 12.50 & 28.38& 11.00& 17.83& 39.33& 21.56& 12.83&       32.97&       14.82&   14.39&     25.53&   14.64\\   
         & Llama 3-instruct (8b) & \textbf{44.98}& \textbf{24.33}& 34.33& 45.19& \textbf{24.56}& 32.17& 55.41& 37.44& 30.33&       \textbf{48.53}&       \textbf{28.78}&   34.65&     42.98&   31.13\\   
         \cmidrule(lr){2-2}\cmidrule(lr){3-5} \cmidrule(lr){6-8} \cmidrule(lr){9-11} \cmidrule(lr){12-16}
         & gpt-3.5-turbo-0301 & 40.49& 20.67& 64.67& \textbf{45.31}& 24.00& 61.17& \textbf{58.59}& 36.22& 58.17& 48.13& 26.96& 61.34& 53.41& 40.71\\
         & gpt-3.5-turbo-0613 & 37.68& 22.22& 80.50& 42.22& 24.44& 76.50& 55.53& 37.67& \textbf{76.00}& 45.14& 28.11&  77.67&   \textbf{58.15}&      \textbf{47.93}\\
         \cmidrule(lr){2-2}\cmidrule(lr){3-5} \cmidrule(lr){6-8} \cmidrule(lr){9-11} \cmidrule(lr){12-16}
         & Aya-23 (8b) & 31.69& 19.89& 82.17& 35.82& 20.89& \textbf{81.67}& 55.30& \textbf{39.78}& 74.00& 40.94& 26.85& \textbf{79.28}& 56.28& 47.82\\
         
         \midrule
         \addlinespace
         \midrule
         
         \multirow{9}{*}{\rotatebox[origin=l]{90}{\textbf{Turkish}}} & \XLMRBase  & 22.65& 14.78&                      68.50& 32.68& 17.44&                      59.33& 41.71& 29.56&          57.17& 32.35& 20.59&    61.67&  44.08&         37.02\\
         & \XLMRDomain & 5.46& 3.22&                      89.33& 11.20& 5.11&                      77.83& 12.40& 6.67&          82.00& 9.69& 5.00&    83.05&   39.05&        36.22\\
         \cmidrule(lr){2-2}\cmidrule(lr){3-5} \cmidrule(lr){6-8} \cmidrule(lr){9-11} \cmidrule(lr){12-16}
         & MAD-X+Domain & 2.15&1.33& 96.00&11.33&6.00&                          90.33&30.97&20.78&              66.17&14.82&9.37&  84.17&  42.64&   39.04\\
         & MAD-X$^2$  &3.97&2.78& \textbf{97.17}& 8.43&4.89&      \textbf{93.17}&21.74&15.89&\textbf{83.50}&11.38&7.85& \textbf{91.28}&   43.34&   41.22\\
         \cmidrule(lr){2-16}
         & Llama 2-chat (13b) & 18.11 &  9.00 & 5.00 & 22.16 & 9.22 & 6.00 & 22.27& 9.00& 4.83&       20.85&       9.07&   5.28&     14.62&   7.55\\   
         & Llama 3-instruct (8b) & \textbf{46.27}& \textbf{31.67}& 25.17& 54.06& 28.56& 36.50& 53.91& 32.67& 26.50&       51.41&       30.97&   29.39&     \textbf{59.35}&   30.34\\   
         \cmidrule(lr){2-2}\cmidrule(lr){3-5} \cmidrule(lr){6-8} \cmidrule(lr){9-11} \cmidrule(lr){12-16}
         & gpt-3.5-turbo-0301 & 36.58& 20.33& 68.33& 53.63& 25.00& 60.83& 53.67& 27.44& 54.50& 47.96& 24.26& 61.22& 53.26& 39.04\\
         & gpt-3.5-turbo-0613 & 44.26& 28.56& 75.17& \textbf{57.29}& \textbf{32.33}& 63.67& \textbf{56.14}& \textbf{33.78}& 57.67& \textbf{52.56}& \textbf{31.56}&  65.50& 57.74& \textbf{45.13}\\
         \cmidrule(lr){2-2}\cmidrule(lr){3-5} \cmidrule(lr){6-8} \cmidrule(lr){9-11} \cmidrule(lr){12-16}
         & Aya-23 (8b) & 41.74& 31.11& 65.83& 52.93& 30.78& 59.83& 52.59& 32.56& 54.33& 49.09& 31.48& 60.00& 53.45& 42.89\\
         
         \midrule
         \addlinespace
         \midrule
         
         \multirow{9}{*}{\rotatebox[origin=l]{90}{\textbf{Chinese}}} & \XLMRBase  & 0.11& 0.11&                      32.33& 0.69& 0.56&                      35.67& 39.67& 24.44&          49.33& 13.49& 8.37&    39.11&  23.74&         20.67\\
         & \XLMRDomain & 0.00& 0.00&                      48.17& 0.28& 0.22&                      62.33& 1.79& 1.00& \textbf{98.00}& 0.69& 0.41&    69.50&   28.21&        28.05\\
         \cmidrule(lr){2-2}\cmidrule(lr){3-5} \cmidrule(lr){6-8} \cmidrule(lr){9-11} \cmidrule(lr){12-16}
         & MAD-X+Domain & 0.00&0.00&                      \textbf{92.00}&0.39&0.33&                          \textbf{85.00}&32.21&20.22&              60.17&10.87&6.85&  \textbf{79.06}&  38.43&   36.02\\
         & MAD-X$^2$  &0.11&0.11&   79.67& 0.17&0.11&      83.67&33.24&22.00&68.67&11.17&7.41&     77.34&   37.64&   35.38\\
         \cmidrule(lr){2-16}
         & Llama 2-chat (13b) & 13.05 & 2.44 & 16.17 & 12.39& 3.89& 17.50&     10.86& 1.89& 14.67&       12.10& 2.74&  16.11&  13.70& 8.09\\   
         & Llama 3-instruct (8b) & \textbf{36.61}& 33.33& 29.17& \textbf{27.52}& 23.33& 32.17& 33.50& 21.00& 26.83&       32.54&       25.89&   29.39&     31.28&   27.29\\   
         \cmidrule(lr){2-2}\cmidrule(lr){3-5} \cmidrule(lr){6-8} \cmidrule(lr){9-11} \cmidrule(lr){12-16}
         & gpt-3.5-turbo-0301 & 27.12& 24.78& 57.00& 19.50& 16.56& 60.50& 18.86& 15.00& 43.50& 21.83& 18.78& 53.67& 34.56& 32.73\\
         & gpt-3.5-turbo-0613 & 35.31& 34.44& 66.50& 26.01& 25.44& 71.67& 27.88& 21.78& 53.83& 29.73& 27.22&  64.00&   43.44&    \textbf{41.93}\\
         \cmidrule(lr){2-2}\cmidrule(lr){3-5} \cmidrule(lr){6-8} \cmidrule(lr){9-11} \cmidrule(lr){12-16}
         & Aya-23 (8b) & 35.44& \textbf{35.44}& 56.83& 26.74& \textbf{26.74}& 65.33& \textbf{50.39}& \textbf{33.67}& 47.33& \textbf{37.52}& \textbf{31.95}& 56.50& \textbf{45.11}& 41.77\\
         \bottomrule

    \end{tabular}
    }
    \vspace{-1mm}
    \caption{Results of the base models and adapter-based methods on the M2QA benchmark using the F1/EM score definitions by SQuAD 2.0 \cite{rajpurkar-etal-2018-know}.
    See our discussion of this metric's potential flaws in Section \ref{sec:ablation:squad_metric_multilingual}.
    For the answerable questions, we report the F1 and Exact Match (EM) scores. For the unanswerable (unansrbl) questions, we only include the F1 score as the EM score is identical to it by definition. The average is taken across datapoints.
    The best score for each language in each column is bold.}
    \vspace{-2.5mm}
    \label{tab:full_results}
\end{table*}

\subsubsection{Performance of Existing Models (RQ1)}
We first investigate how well existing models perform on the dataset. This includes LLMs and fine-tuned XLM-R baselines. We observe that out of all the approaches we evaluated, \texttt{gpt-3.5-turbo-0613} performs best with an average F1 score of 53.11 followed by Aya 23 with 51.61, \texttt{gpt-3.5-turbo-0301} with 47.08, Llama 3-instruct with 44.54. 

Llama 2-chat (13b), with an average F1 score of $17.95$, performs poorly in Turkish and Chinese, especially on the answerable questions.
This is not surprising considering that LLama-2 is trained mainly on English text.
In the zero-shot setting, Llama-2 often produces long responses that mix English with the target language. However, these issues are less pronounced in German, leading to comparatively better performance.
To improve the performance of LLMs, we investigated using few-shot prompts. As detailed in Appendix \ref{sec:appendix:five-shot-llm-results}, only Llama 2 and \texttt{gpt-3.5-turbo-0301} consistently benefitted from this.

\XLMRBase has an average F1 score of 37.73 and performs well across languages, despite being smaller. \XLMRDomain, with an average F1 score of 36.36, performs particularly poorly on answerable questions.
This indicates that performing intermediate fine-tuning on English domain data is not only insufficient for domain transfer but actually hurts the performance, at least for German and Turkish. This is potentially caused by catastrophic forgetting of language-specific information \cite{French1999CatastrophicFI}.

To gain further insights into the performance of GPT-3.5, we manually inspected German questions for which all four GPT-3.5 setups achieved an F1 score lower than 25, which are 942 questions in total, or 20.9\% of the German QA instances. We randomly sampled 50 questions from this subset to analyze the responses of the GPT-3.5 models.
We found that in 72\% of the cases, the question and answer are correctly annotated in the data, but the model either makes erroneous predictions (58\%) or generates a correct answer instead of extracting it (14\%). We further identified issues with inconsistent annotations (22\%, i.e. 4.6\% of all German data), questions with multiple plausible answers (4\%), and the evaluation metric (2\%).
We detail this investigation in Appendix \ref{sec:appendix-gpt-answer-investigation}.

\subsubsection{Hard Domains and Languages (RQ2)}
We now explore which languages and domains are particularly hard to tackle for the existing models. As per Table \ref{tab:full_results}, for all explored models, the scores in German and Turkish are notably higher than the scores in Chinese, suggesting that this transfer is harder for the models. We revisit this observation in Section \ref{sec:ablation}.
Moreover, the performance in the news domain is higher than in creative writing and reviews. This shows that the model's domain transfer abilities still have room for improvement.

Performance on creative writing and product reviews varies by language. For German and Turkish, the results on product reviews are considerably better than on creative writing on the answerable questions, whereas in Chinese, the results are considerably better in creative writing for GPT-3.5 and Llama.
This highlights the need for a joint evaluation of language and domain transfer. To investigate isolated cross-lingual and cross-domain transfer, we evaluated further setups, but could not find improved performance (Appendix \ref{sec:appendix:isolated-domain-and-language-transfer}).

\begin{table*}[t]
    \centering
    \small
    \begin{tabular}{lccc}
        \toprule
        & Creative Writing & Product Reviews& News \\
        & F1 answerable& F1 answerable & F1 answerable \\
        \midrule
        \XLMRBase  & 8.78 (+8.67)& 8.82 (+8.13)& 41.03 (+1.36)\\
        \XLMRDomain & 7.24 (+7.24)& 4.87 (+4.59)& 1.94 (+0.15)\\
        \cmidrule(lr){1-1} \cmidrule(lr){2-4}
        MAD-X+Domain & 0.93 (+0.93)& 2.73 (+2.34)& 33.33 (+1.12)\\
        MAD-X$^2$    & 4.32 (+4.21)& 2.96 (+2.79)& 33.57 (+0.33)\\
        \cmidrule(lr){1-4}
        Llama 2-chat (13b) & 18.52 (+5.47)& 19.49 (+7.10)& 28.97 (+18.11)\\
        Llama 3-instruct (8b) & 51.35 (+14.74)& 43.07 (+15.55)& 48.49 (+14.99)\\
        \cmidrule(lr){1-1} \cmidrule(lr){2-4}
        gpt-3.5-turbo-0301 & 45.96 (+18.84)& 40.21 (+20.71)& 46.87 (+28.01)\\
        gpt-3.5-turbo-0613 & 48.22 (+12.91)& 40.54 (+14.53)& 49.27 (+21.93)\\
        \cmidrule(lr){1-1} \cmidrule(lr){2-4}
        Aya-23 (8b)         & 50.30 (+14.86)& 40.64 (+13.90)& 55.84 (+5.45)\\
        \bottomrule
    \end{tabular}
    \vspace{-1mm}
    \caption{Chinese results using the adapted SQuAD 2.0 metric with word tokenization instead of whitespace tokenization, affecting F1 scores on answerable questions. Relative changes to Table \ref{tab:full_results} are shown in parentheses. LLMs use zero-shot prompts.}
    \vspace{-2.5mm}
    \label{tab:jieba_tokenization}
\end{table*}

\subsubsection{Modular setups (RQ3)}
Finally, we use M2QA to evaluate our two modular adaptation setups: \MADXDomain and \mbox{\MADXSquare}. Based on our results (Table \ref{tab:full_results}), these setups achieve average scores on par with \XLMRBase in German and Turkish, and notably improve the Chinese score. We note that this increase primarily stems from the improved performance on unanswerable questions, while the performance on answerable questions declines. 
Despite similar overall performance between \MADXDomain and \MADXSquare, a notable difference lies in the number of update steps during training: \MADXDomain was trained a total of 1M training steps (100k for each domain and 250k for each language, with a batch size of 64), while \MADXSquare only needs 250k training steps with batch size 16 to achieve similar performance. This highlights \MADXSquare computational efficiency, indicating the potential for simultaneous training of language and domain adapters.

\section{Further Analysis}
\label{sec:ablation}
In contrast to English, German, and Turkish, which use whitespace characters to separate words, in Chinese typesetting the use of whitespace is not \textit{required}. While the texts from our Chinese product review and creative writing sources do not contain whitespaces, Chinese news do. We hypothesize that this typographical difference between Chinese and the other languages can lead to a substantial drop in measured performance (e.g. \XLMRBase achieves an F1 score of $0.11$ on answerable creative writing questions), and investigate this further.

\subsection{SQuAD Metric - Adaptation for Chinese}
\label{sec:ablation:squad_metric_multilingual}
For the evaluation in Section \ref{sec:results}, we have used the F1/EM definitions of SQuAD 2.0, which is widely adopted and has been previously used to evaluate multilingual extractive QA \cite[e.g.][]{artetxe-etal-2020-cross}. During the metric calculation, this implementation splits words by whitespace -- however, if whitespaces are not available, the whole text is considered as one long token, rendering the rest of the calculation invalid. We modify the implementation to make the metric applicable to Chinese texts without whitespace tokenization by splitting the text into tokens using the off-the-shelf jieba tokenizer\footnote{\url{https://github.com/fxsjy/jieba} v0.42.1}.
The resulting measurements, shown in Table \ref{tab:jieba_tokenization}, differ substantially from those in Table \ref{tab:full_results}, suggesting that the SQuAD metric implementation needs adjustment for multilingual extractive QA evaluation.
Even for texts from the news domain which contain whitespace, the tokenizer-based version of the metric results in higher scores. The tokenizer splits the Chinese text into smaller tokens than whitespace tokenization, allowing a finer-grained score. Moreover, the XLM-R-based methods struggle to make meaningful predictions for text without whitespace (see Section \ref{sec:whitespace}). Since the score only improves for spans close to the gold span, the improvement for LLMs is bigger than for XLM-R-based methods.

\begin{table*}[t]
    \centering
    \small
    \begin{tabular}{l ccccccccc}
    \toprule
     & \multicolumn{3}{c}{\textbf{Creative Writing}} & \multicolumn{3}{c}{\textbf{Product Reviews}}  & \multicolumn{3}{c}{\textbf{News}} \\
     & \multicolumn{2}{c}{answerable} & unansrbl& \multicolumn{2}{c}{answerable} & unansrbl& \multicolumn{2}{c}{answerable} & unansrbl\\
      & F1    & EM & F1   & F1    & EM & F1    & F1    & EM & F1\\
    \midrule
    original text &  8.78 & 0.11 & 32.33 & 8.82 & 0.56& 35.67 & 41.03 &  24.44 & 49.33\\
    + jieba whitespaces  & 25.24 & 16.89 & 70.00 & 22.11 & 12.89 & 60.00 & 28.36 & 7.44 & 50.50 \\
    \bottomrule
    \end{tabular}
    \vspace{-1mm}
    \caption{Results of \XLMRBase on the original texts and with added whitespace, evaluated with the adapted SQuAD metric using a word tokenizer instead of whitespace tokenization. }
    \vspace{-2.5mm}
    \label{table:abalation_whitespaces}
\end{table*}

\subsection{Adding Whitespaces to Chinese Text}
\label{sec:whitespace}
Having examined the predictions of the XLM-R-based methods, we found that training on English SQuAD data leads to XLM-R returning spans surrounded by whitespace as answers. If the Chinese text does not contain whitespaces, XLM-R-based methods either classify the question as unanswerable or return the whole passage as the answer. 
To explore the impact of this issue, we re-run the \XLMRBase setup but added whitespace to the texts between jieba-determined words. The results in Table \ref{table:abalation_whitespaces} show that this modification leads to improved performance on Chinese texts with no whitespace (+24.9 F1 points for creative writing, +17.7 F1 points for product reviews) but reduces the measured performance on texts with whitespace (-7.1 F1 points for news).

The improved performance on texts that previously had no whitespace suggests that language transfer methods like MAD-X struggle to transfer tasks to languages without inherent whitespace. The reduced performance on texts that already contained whitespace indicates that adding whitespace between jieba-determined words is not yet optimal.
This suggests that typographical features of the source data can affect measured performance and should be taken into account when experimenting with non-Latin-based languages. Heuristics, i.e. whitespaces added through tokenization, can help improve performance.

\section{Discussion and Future Work}
M2QA allows us to evaluate joint language and domain transfer across different language models and adaptation approaches.
Our results indicate room for improvement, especially when comparing the results of XLM-R-based models and LLMs.
Since 40\% of M2QA's questions are unanswerable, a naive model that classifies all questions as unanswerable would reach an F1/EM score of 40.0/40.0.
For Chinese, only \texttt{gpt-3.5-turbo-0613} and Aya 23 perform better than this naive strategy, emphasizing the need for more sophisticated domain and language transfer methods. We hope that our resource enables and encourages work on systematically exploring prompts and setups that perform transfer learning across multiple dimensions of language variation.
Future efforts should also aim to add more languages and domains to M2QA, especially for low-resource languages and domains.
We hope that our published annotation protocols and software will facilitate this work.\footnote{The code for all experiments, including hyperparameters, prompts, and the implementation of the annotation environment, is available in our GitHub repository: \url{https://github.com/UKPLab/m2qa}}
Finally, establishing human performance baselines would help us understand how far NLP systems are from achieving human-level extractive QA performance across languages and domains.

\section{Conclusion}
Generalization is a central goal of NLP that is yet unsolved. Language and domain are two main axes of variation for natural languages -- yet the lack of cross-lingual cross-domain datasets has prevented systematic evaluation of NLP models and transfer approaches across languages and domains. To address this, we introduce M2QA, a multi-domain multilingual question answering benchmark with over 13k human-annotated instances across three typologically diverse languages (German, Turkish, Chinese) and three distinct domains (product reviews, news, creative writing). 
Our evaluation includes XLM-R baselines, LLMs (GPT-3.5, Aya 23, Llama 2 and 3), and adapter-based setups (\MADXDomain and \MADXSquare), revealing a large gap between LLMs and fine-tuned LMs. We expect that M2QA will help close this gap, increase our understanding of generalization, and find more effective domain and language transfer methods.

\section{Limitations}
A major obstacle to including more languages and domains into M2QA has been a \emph{severe shortage} of \emph{clearly and openly licensed} unlabeled texts in under-represented language-domain combinations -- due to the restrictive copyright in many domains (news, books), and due to the lack of explicit licensing practices in others. While we made an effort to diversify the selection of languages and domains in M2QA, the dataset only covers a small subset of all existing languages and domains.
Potential solutions to this could be to clarify or obtain a license for research use from the owners of the textual data, as well as to experiment with data synthetically generated via paraphrasing or machine translation.
However, translations introduce considerable issues, including lowered lexical diversity, "translationese", and lack of cultural idiosyncrasies, as discussed in the introduction and Section \ref{sec:m2qa_dataset_requirements}.
This exploration, as well as the comparison between the results on synthetic and natural QA data, is left to the future.

Since some of the data sources in M2QA are widely used (e.g. Gutenberg Corpus or Amazon Reviews), there is a risk that LLMs have observed some of the unlabeled data during their pre-training. The unavailability of pre-training data for LLaMa 2, Llama 3, GPT-3.5 and Aya 23 prevents us from investigating whether this is the case. To prevent contamination of future experimental setups with the labelled data, we employ protective measures, following \citet{jacovi-etal-2023-stop}: We release the data in encrypted form with a CC-BY-ND 4.0 license.\footnote{\url{https://creativecommons.org/licenses/by-nd/4.0/legalcode}}

We evaluate with XLM-R for a consistent setup with MAD-X \cite{pfeiffer-etal-2020-mad}.

\section*{Acknowledgments}
We gratefully acknowledge the support of Microsoft with a grant for access to OpenAI GPT models via the Azure cloud (Accelerate Foundation Model Academic Research). Co-funded by the LOEWE Distinguished Chair “Ubiquitous Knowledge Processing”, LOEWE initiative, Hesse, Germany (Grant Number: LOEWE/4a//519/05/00.002(0002)/81) and by the European Union (ERC, InterText, 101054961).  Views and opinions expressed are however those of the author(s) only and do not necessarily reflect those of the European Union or the European Research Council. Neither the European Union nor the granting authority can be held responsible for them. Hannah Sterz thanks the Cambridge Trust for their support via the International Scholarship. We thank Sebastian Ruder for insightful feedback and Massimo Nicosia, Yongxin Huang and Thy Thy Tran for helpful comments on an earlier draft of this paper. Finally, we thank Furkan Şahinuç and Qian Ruan for their help in validating multilingual data.

\bibliography{anthology,custom}
\bibliographystyle{acl_natbib}

\appendix
\section{Passage Curation}
\label{sec:appendix-passage-curation}
Table \ref{tab:datasets} shows the datasets we selected and their licensing information. The final dataset should contain 300 passages for each language and domain combination. To preprocess the data and prepare the passages, we need to distinguish between the domains that have multiple passages per document and the ones with one passage per document. For ones where a document is one passage, we filter out documents that are too short and too long. From the remaining documents, we randomly sample 300. For the domains that have multiple passages per document, we first exclude the ones that are too short to feature at least three passages. Then, we sample documents and split them into passages using the WTP segmentation model \cite{minixhofer-etal-2023-wheres}. We use only documents with at least three passages. The German creative writing of the Gutenberg corpus required a different setup. Because of the different formatting of footnotes, references, and diverse formatting of bold, underlined, and cursive text, we manually extracted 300 passages from 6 fiction creative writing that had licenses that made them free to use. The passages from all domains are then stripped of newline characters, tabs, and multiple consecutive white spaces. The creative writing passages for Turkish and Chinese are taken from an online social reading platform where people can publish their own stories. We select texts published in the public domain or with a Creative Common License. To ensure that no author's notes or unsuitable or offensive texts, such as comments or sensitive topics, are in the passage, we manually check the translated \footnote{We use DeepL for translation.} passages.

\begin{table*}[]
    \begin{threeparttable}
        \centering
        \small
        \begin{tabularx}{\textwidth}{llcXX}
        \toprule
            Language & Domain & Multiple Passages & Datasource & License \\
            \midrule
            \mr{5}{German}  & product reviews   & no    & Amazon Reviews \cite{keung-etal-2020-multilingual}    & Usage permitted by Amazon for academic research\tnote{1}. \\ \cmidrule{2-5}
                            & news              & yes   & 10kGNAD\tnote{2}                      & CC BY-NC-SA 4.0                                           \\ \cmidrule{2-5}
                            & creative writing  & yes   & Gutenberg Corpus \cite{Gerlach2018ASP}& Manually selected text passages from open-license books. \\
            \midrule
            \mr{5}{Turkish} & product reviews   & no    & Turkish product reviews\tnote{3}      & CC BY-SA 4.0                                              \\ \cmidrule{2-5}
                            & news              & yes   & BilCat \cite{Toraman2011Developing}   & MIT License                                               \\ \cmidrule{2-5}
                            & creative writing  & yes   & Wattpad\tnote{4}                      & Manually selected text passages from Creative Commons or Public Domain publications. \\
            \midrule
            \mr{6}{Chinese} & product reviews   & no    & Amazon Reviews \cite{keung-etal-2020-multilingual}    & Usage permitted by Amazon for academic research\tnote{1}. \\ \cmidrule{2-5}
                            & news              & yes   & CNewSum \cite{Wang2021CNewSum}        & MIT License                                               \\ \cmidrule{2-5}
                            & creative writing  & yes   & Wattpad\tnote{4}                      & Manually selected text passages from Creative Commons or Public Domain publications. \\
        \bottomrule
        \end{tabularx}
        \vspace{-2mm}
        \caption{The original datasets used for annotation}
        \label{tab:datasets}
        \begin{tablenotes}
            \footnotesize
            \item [1] \url{https://github.com/awslabs/open-data-docs/blob/main/docs/amazon-reviews-ml/license.txt}
            \item [2] \url{https://github.com/tblock/10kGNAD} using the One Million Posts dataset by \citet{schabus2017one}
            \item [3] \url{https://huggingface.co/datasets/turkish_product_reviews}
            \item [4] \url{https://www.wattpad.com/}
        \end{tablenotes}
    \end{threeparttable}
\end{table*}

\subsection{Lexical Diversity of the Domains}
\label{sec:appendix:passages_lexical_diversity}
\begin{figure*}[]
    \centering
     \begin{subfigure}[b]{0.245\textwidth}
         \centering
         \includegraphics[width=\textwidth]{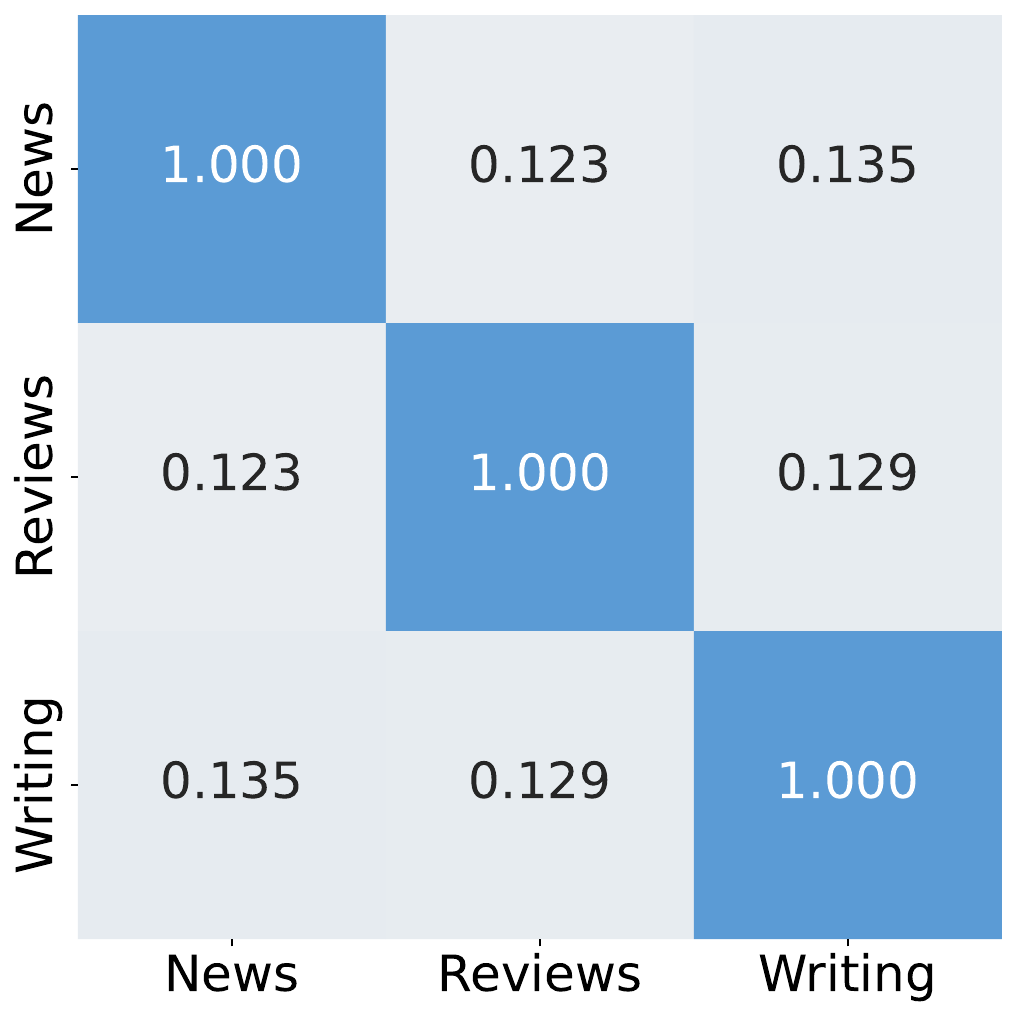}
         \caption{German}
     \end{subfigure}
     \hfill
     \begin{subfigure}[b]{0.245\textwidth}
         \centering
         \includegraphics[width=\textwidth]{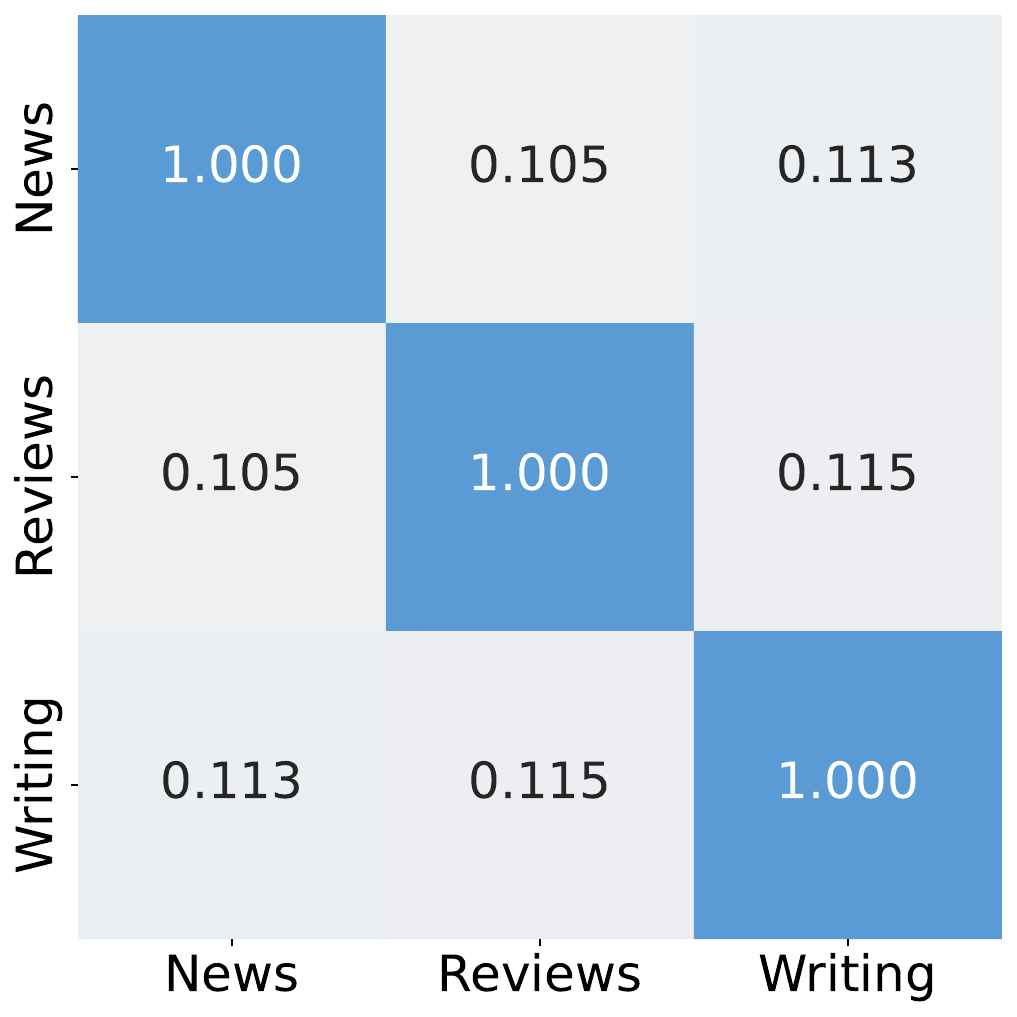}
         \caption{Turkish}
     \end{subfigure}
     \hfill
     \begin{subfigure}[b]{0.245\textwidth}
         \centering
         \includegraphics[width=\textwidth]{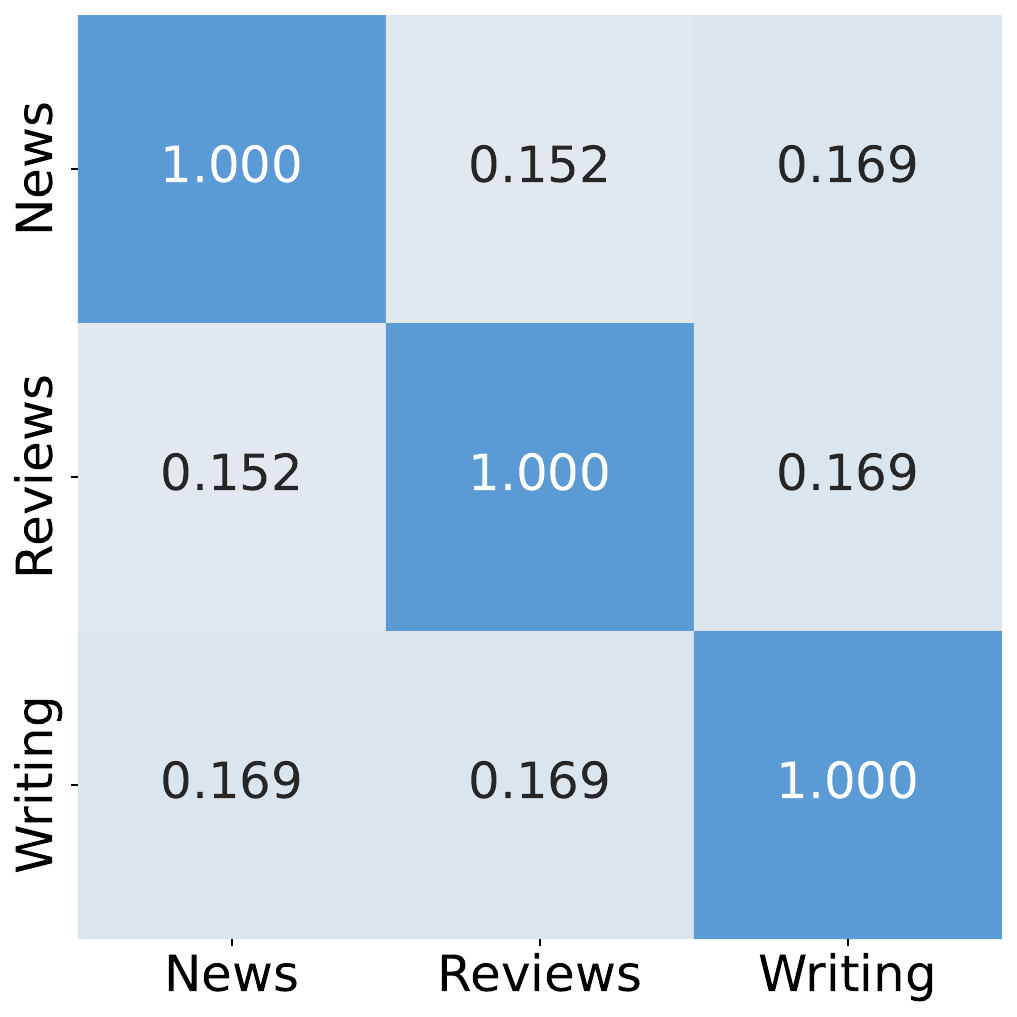}
         \caption{Chinese}
     \end{subfigure}
        \caption{Jaccard similarity coefficient of the vocabularies between the domains, per language.}
        \label{fig:vocab}
        \vspace{-3.3mm}
\end{figure*}

To quantify lexical diversity in the data, in Figure \ref{fig:vocab}, we report the Jaccard similarity coefficient of the vocabularies between the different domains in one language.\footnote{We use nltk \url{https://www.nltk.org} for German and Turkish tokenization, and jieba \url{https://github.com/fxsjy/jieba} for Chinese} As we observe, the domains in M2QA indeed show low vocabulary overlaps, making our dataset a challenging target for domain adaptation across languages.

\section{Baseline Training}
\label{sec:appendix-baseline-training}

All our models are based on XLM-R-base \cite{conneau-etal-2020-unsupervised}, a multilingual 270M parameter model. \XLMRBase is directly fine-tuned on SQuAD 2.0, while \XLMRDomain has been domain-adapted prior to fine-tuning on SQuAD 2.0 \cite{rajpurkar-etal-2018-know}. Every not-mentioned hyperparameter is the default parameter of Hugging Face Transformers \cite{wolf-etal-2020-transformers} version 4.26.1.

\paragraph{\XLMRBase}
We train XLM-R on SQuAD 2.0 for 100,000 update steps, use early stopping with patience of 5 and evaluate every 1000 steps. We use a batch size of 64, 1000 warmup steps, a learning rate of 1e-4, linear learning rate decay and a sequence length of 512.

\paragraph{\XLMRDomain}
We first train an individual model for each domain via MLM on the data displayed in Table \ref{table:appendix:english-domain-datasources}. We train for 100,000 update steps with a batch size of 16, a learning rate of 1e-4, linear learning rate decay and a sequence length of 512. Following \citet{wettig-etal-2023-mask}, we use an MLM probability of 40\% since XLM-R-base has a comparable size to bert-large.
After this training, we fine-tune every domain-adapted XLM-R model on SQuAD 2.0 with the same parameters used for \XLMRBase.

\begin{table}[]
    \centering
    \small
    \begin{tabular}{lc}
        \toprule
        Domain & Datasource\\
        \midrule
        Wikipedia & Wikipedia \cite{wikidump} \\
        Creative Writing& bookcorpus \cite{zhu2015aligning}\\
        Product Reviews & Amazon Reviews \cite{keung-etal-2020-multilingual}\\
        News & CNN Dailymail \cite{hermann2015teaching} \\
        \bottomrule
    \end{tabular}
    \vspace{-0.5em}
    \caption{Texts used for adapting the domain of \XLMRDomain and the domain adapters of MAD-X+Domain.}
    \vspace{-1.2em}
    \label{table:appendix:english-domain-datasources}
\end{table}

\section{MAD-X+Domain \& MAD-X$^2$ Training}
\label{sec:appendix-mad-xexperiments-traininig}

We use the \textit{Adapters} library \cite{poth-etal-2023-adapters} for the adapter and model implementations. 

\begin{figure}
    \centering
     \begin{subfigure}[b]{0.45\textwidth}
         \centering
         \includegraphics[width=\textwidth]{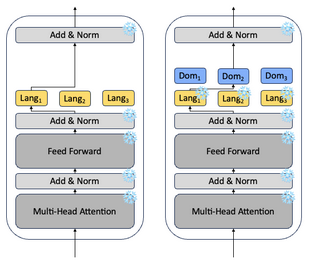}
         \caption{\MADXDomain: The language adapter is trained as proposed in the MAD-X setup, and the domain adapter is trained in a second step with the corresponding frozen language adapter activated.}
     \end{subfigure}
     \hfill
     \begin{subfigure}[b]{0.45\textwidth}
         \centering
         \includegraphics[width=0.5\textwidth]{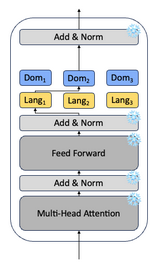}
         \caption{\MADXSquare: During training, the language and domain adapters are trained simultaneously, and each training sample is routed through the corresponding language and domain adapter. }
     \end{subfigure}\hfill     
    \caption{The different pre-training setups for \MADXDomain, and \MADXSquare}
    \label{fig:appendix:madXsetups_frozen_adapters}
\end{figure}

\paragraph{MAD-X+Domain}
The domain adapters are Pfeiffer Bottleneck Adapters with a reduction factor of 2.
We train each domain adapter for 100,000 update steps with a batch size of 16, 1000 warmup steps, learning rate of 1e-4 and linear learning rate decay via masked language modelling on English data. The data sources used for each adapter are listed in Table \ref{table:appendix:english-domain-datasources}. Since we train on English data, we activate the English MAD-X \cite{pfeiffer-etal-2020-mad} adapter. Following \citet{wettig-etal-2023-mask}, we use an MLM probability of 40\% since XLM-R-base has a comparable size to bert-large.
Overall, the domain and language adapters cumulatively used 1,400,000 update steps with a batch size of 64 (the four language adapters were trained with 250,000 steps, the four domain adapters with 100,000).

The QA head adapter, also a Pfeiffer Bottleneck Adapter with a reduction factor of 2, was trained on SQuADv2 \cite{rajpurkar-etal-2018-know} using the same hyper-parameters used for fine-tuning \XLMRBase. Since SQuADv2 is based on English Wikipedia text passages, we activated the English and Wikipedia adapter during training.

\paragraph{MAD-X$^2$}
We use the same hyper-parameters as for the training of the domain adapters of \MADXDomain. The only parameter changed is the number of update steps where we train every domain and language adapter for 62,500 update steps, resulting in a total of 250,000 update steps. The corpora used for the MLM training are listed in Table \ref{table:appendix:mad-x-2-data} along with the number of steps trained on each corpus. Due to the absence of open-license text corpora, we do not train on Chinese and Turkish creative writing corpora. The QA head adapter is trained afterwards identical to the QA head adapter of \MADXDomain.

\begin{table*}[]
    \begin{threeparttable}
    \centering
    \small
    \begin{tabularx}{\textwidth}{llYc}
        \toprule
        Language& Domain& Datasource& steps trained \\
        \midrule
        \mr{4}{English}& Wikipedia& Wikipedia \cite{wikidump} & 10417\\
            & Creative Writing& PG-19 \cite{rae2019compressive}  & 31250\\
            & Product Reviews& Amazon Reviews \cite{keung-etal-2020-multilingual} & 10416\\
        & News& CNN Dailymail \cite{hermann2015teaching} & 10417\\
        \midrule
        \mr{5}{German}& Wikipedia& Wikipedia \cite{wikidump} & 10417\\
        & Creative Writing& Opus Books \cite{tiedemann2012parallel} \& Corpus of German-Language Fiction \cite{fischer2015does} & 31250\\
        & Product Reviews& Amazon Reviews \cite{keung-etal-2020-multilingual} & 10416\\
        & News& MLSUM \cite{scialom-etal-2020-mlsum} & 10417\\
        \midrule
        \mr{4}{Turkish}& Wikipedia& Wikipedia \cite{wikidump} & 20833\\
        & Creative Writing&  & 0\\
        & Product Reviews&  Turkish Product Reviews\tnote{1}& 20833\\
        & News&  XL-Sum \cite{hasan-etal-2021-xl}& 20833\\
        \midrule
        \mr{4}{Chinese}& Wikipedia& Wikipedia \cite{wikidump} & 20833\\
        & Creative Writing&  & 0\\
        & Product Reviews& Amazon Reviews \cite{keung-etal-2020-multilingual} & 20833\\
        & News&  XL-Sum \cite{hasan-etal-2021-xl}& 20833\\        
        \bottomrule
    \end{tabularx}
    \vspace{-2mm}
    \caption{MAD-X² training data sources \& number of steps trained}
    \label{table:appendix:mad-x-2-data}
    \begin{tablenotes}
        \footnotesize
        \item [1] \url{https://huggingface.co/datasets/turkish_product_reviews}
    \end{tablenotes}
    \end{threeparttable}
    \vspace{-4mm}
\end{table*}

\section{LLM Evaluation}

\subsection{Five-Shot LLM Results}
\label{sec:appendix:five-shot-llm-results}
We explored if providing a few-shot prompt could enhance the performance of the LLMs. In Table \ref{tab:full_results}, we present the results of the LLMs using a five-shot prompt. Upon comparing these results with the zero-shot results in Table \ref{tab:appendix:five_shot_results}, we see that the five-shot prompt does not consistently improve performance.
The only models benefitting from the five-shot prompt are Llama 2 and \texttt{gpt-3.5-turbo-0301}.

\begin{table*}[t]
    \centering
    \resizebox{\textwidth}{!}{%
    \begin{tabular}{lllcccccccccccccc}
        \toprule
         &  & & \multicolumn{3}{c}{\textbf{Creative Writing}} & \multicolumn{3}{c}{\textbf{Product Reviews}}  & \multicolumn{3}{c}{\textbf{News}} & \multicolumn{5}{c}{\textbf{Average}}   \\
                   & & & \multicolumn{2}{c}{answerable} & unansrbl& \multicolumn{2}{c}{answerable} & unansrbl& \multicolumn{2}{c}{answerable} & unansrbl& \multicolumn{2}{c}{answerable} & unansrbl & \multicolumn{2}{c}{Total Average}\\
         & Model &      & F1    & EM & F1   & F1    & EM & F1    & F1    & EM & F1    & F1    & EM & F1    & F1 & EM\\
         \midrule

         \multirow{5}{*}{\rotatebox[origin=l]{90}{\textbf{German}}} 
          & Llama 2-chat (13b) &(5-shot) &    22.61&               12.33&              75.17&             20.52&     12.67&                  75.33&       29.33&        19.89&        77.33&       24.15&       14.96&   75.94&     44.87 (+19.34)&   39.36 (+24.72)\\
         & Llama 3-instruct (8b) &(5-shot) &    \textbf{43.44}&               \textbf{23.89}&              35.50&             39.82&     20.56&                  25.50&       54.71&        35.56&        28.00&       45.99&       26.67&   29.67&     39.46 (-3.52)&   27.87 (-3.26)\\
         \cmidrule(lr){2-3}\cmidrule(lr){4-6} \cmidrule(lr){7-9} \cmidrule(lr){10-12} \cmidrule(lr){13-17}
         & gpt-3.5-turbo-0301 &(5-shot)& 40.00& 21.89& 76.83& \textbf{47.81}& \textbf{24.33}& 60.67& \textbf{61.18}& 38.22& 59.67& \textbf{49.66}& 28.15&  65.72& 56.09 (+2.68)&43.18 (+2.47)\\
         & gpt-3.5-turbo-0613 &(5-shot) & 34.97& 22.67& 83.33& 40.36&23.33& 79.17&58.50&\textbf{39.44}&70.83&44.61& \textbf{28.48}&    77.78&   \textbf{57.88} (-0.27)&    \textbf{48.20} (+0.27)\\
         \cmidrule(lr){2-3}\cmidrule(lr){4-6} \cmidrule(lr){7-9} \cmidrule(lr){10-12} \cmidrule(lr){13-17}
         & Aya-23 (8b) &(5-shot)& 16.44& 10.78& \textbf{90.83}& 19.85& 12.11& \textbf{93.67}& 38.00& 28.44& \textbf{89.33}& 24.76& 17.11&  \textbf{91.28}& 51.37 (-4.91)&46.78 (-1.04)\\
         
         \midrule
         \addlinespace
         \midrule
         
         \multirow{5}{*}{\rotatebox[origin=l]{90}{\textbf{Turkish}}} 
         & Llama 2-chat (13b) &(5-shot) &    7.18&               5.22&              \textbf{84.83}&             8.48&     4.78&                  \textbf{91.00}&       9.78&        6.33&        \textbf{88.17}&       8.48&       5.44&   \textbf{88.00}&     40.29 (+25.67)&   38.47 (+30.92)\\
         & Llama 3-instruct (8b) &(5-shot) &    \textbf{38.73}&               \textbf{24.33}&              47.33&             41.27&     21.89&                  42.33&       51.15&        30.22&        25.83&       43.72&       25.48&   38.50&     41.63 (-17.72)&   30.69 (+0.35)\\
         \cmidrule(lr){2-3}\cmidrule(lr){4-6} \cmidrule(lr){7-9} \cmidrule(lr){10-12} \cmidrule(lr){13-17}
         & gpt-3.5-turbo-0301 &(5-shot)& 36.82& 22.78& 73.83& \textbf{53.89}& 24.33& 61.67& \textbf{58.13}& \textbf{32.33}& 55.00& \textbf{49.61}& \textbf{26.48}&  63.50& 55.17 (+1.91)&41.29 (+2.25)\\
         & gpt-3.5-turbo-0613 &(5-shot) & 33.58& 20.89& 84.50& 49.98&\textbf{25.78}& 71.17&55.21&32.00&58.00&46.26&26.22&    71.22&   \textbf{56.24} (-1.5)&    44.22 (-0.91)\\
         \cmidrule(lr){2-3}\cmidrule(lr){4-6} \cmidrule(lr){7-9} \cmidrule(lr){10-12} \cmidrule(lr){13-17}
         & Aya-23 (8b) &(5-shot)& 28.52& 21.89& 79.33& 33.14& 20.56& 84.67& 32.95& 20.44& 74.00& 31.54& 20.96&  79.33& 50.66 (-2.79)&\textbf{44.31} (+1.42)\\

         \midrule
         \addlinespace
         \midrule

         \multirow{5}{*}{\rotatebox[origin=l]{90}{\textbf{Chinese}}} 
         & Llama 2-chat (13b) &(5-shot) &    0.71&               0.67& \textbf{95.33}&             1.50&     1.44& \textbf{90.83}&       1.19&        0.78&        \textbf{96.00}&       1.13&       0.96&   \textbf{94.05}&     38.30 (+24.6)&   38.20 (+30.11)\\
         & Llama 3-instruct (8b) &(5-shot) &    \textbf{34.78}&               \textbf{32.22}&              34.83&             21.16&     17.67&                  32.67&       33.21&        20.67&        22.83&       29.72&       23.52&   30.11&     29.88 (-1.4)&   26.16 (--1.13)\\
         \cmidrule(lr){2-3}\cmidrule(lr){4-6} \cmidrule(lr){7-9} \cmidrule(lr){10-12} \cmidrule(lr){13-17}
         & gpt-3.5-turbo-0301 &(5-shot)& 29.19& 26.22& 63.17& 21.05& 16.89& 63.67& 21.55& 17.33& 48.50& 23.93& 20.15&  58.45& 37.74 (+3.18)&35.47 (+2.74)\\
         & gpt-3.5-turbo-0613 &(5-shot) & 29.30& 28.33& 75.83& 16.07&14.67& 82.33&25.44&20.11&57.83&23.60&21.04&    72.00&   42.96 (-0.48)&    \textbf{41.42} (-0.51)\\
         \cmidrule(lr){2-3}\cmidrule(lr){4-6} \cmidrule(lr){7-9} \cmidrule(lr){10-12} \cmidrule(lr){13-17}
         & Aya-23 (8b) &(5-shot)& 28.11& 28.11& 60.33& \textbf{22.74}& \textbf{22.67}& 61.17& \textbf{50.40}& \textbf{34.33}& 50.83& \textbf{33.75}& \textbf{28.37}&  57.44& \textbf{43.23} (-1.88)&40.00 (-1.77)\\         
         \bottomrule

    \end{tabular}
    }
    \caption{Results of the LLMs with five-shot prompts on the M2QA benchmark using the same scores as Table \ref{tab:full_results}. Relative changes to Table \ref{tab:full_results} are shown in parentheses of the \textit{Total Average} column.
    For the answerable questions, we report the F1 and Exact Match (EM) scores. For the unanswerable (unansrbl) questions, we only include the F1 score as the EM score is identical to it by definition. The average is taken across datapoints. The best score for each language in each column is bold.}
    \label{tab:appendix:five_shot_results}
\end{table*}

\subsection{LLM Prompts}
\label{sec:appendix-prompt-details}
Based on the extractive question answering prompt of \citet{lai-etal-2023-chatgpt}, we evaluate the LLMs in a zero-shot and five-shot setting. The zero-shot prompt is displayed in Figure \ref{fig:appendix:zero_shot_prompt_english}. The five-shot prompts contain three answerable and two unanswerable examples. Following \citet{Brown2020Language}, we provide the five examples in a single user prompt, as shown in Figure \ref{fig:appendix:five_shot_prompt_english}.
However, this setup yielded scores close to zero for Llama 2-chat 13B. Hence, we changed the Llama 2-chat 13B setup by providing examples not in a single prompt but as part of the chat history. 

\begin{figure}[]
    \small
    \begin{mdframed}[backgroundcolor=gray!10, linecolor=black, linewidth=1pt, innerleftmargin=10pt, innerrightmargin=10pt, innertopmargin=10pt, innerbottommargin=10pt, leftmargin=0pt, rightmargin=0pt]
        \textbf{System Prompt:} \\
        Task Description: Answer the question from the given passage. Your answer should be directly extracted from the passage, and it should be a single entity, name, or number, not a sentence. If the passage doesn't contain a suitable answer, please respond with 'unanswerable'.
        
        \vspace{5pt}
        \textbf{User:} \\
        Passage: \{CONTEXT\} \\
        Question: \{QUESTION\}\\
        Note: Your answer should be directly extracted from the passage and be a single entity, name, or number, not a sentence. If the passage doesn't contain a suitable answer, respond with 'unanswerable'.\\
        Answer: 
    \end{mdframed}
    \vspace{-3mm}
    \caption{Zero-shot English Prompt}
    \vspace{-2em}
    \label{fig:appendix:zero_shot_prompt_english}
\end{figure}

\begin{figure*}[]
    \small
    \begin{mdframed}[backgroundcolor=gray!10, linecolor=black, linewidth=1pt, innerleftmargin=10pt, innerrightmargin=10pt, innertopmargin=10pt, innerbottommargin=10pt, leftmargin=0pt, rightmargin=0pt]
        \textbf{System Prompt:}\\
        Task Description: Answer the question from the given passage. Your answer should be directly extracted from the passage, and it should be a single entity, name, or number, not a sentence. If the passage doesn't contain a suitable answer, please respond with 'unanswerable'.

        \vspace{5pt}
        \textbf{User:}\\
        Passage: In 2007, BSkyB and Virgin Media became involved in a dispute over the carriage of Sky channels on cable TV. The failure to renew the existing carriage agreements negotiated with NTL and Telewest resulted in Virgin Media removing the basic channels from the network on 1 March 2007. Virgin Media claimed that BSkyB had substantially increased the asking price for the channels, a claim which BSkyB denied, on the basis that their new deal offered "substantially more value" by including HD channels and Video On Demand content which was not previously carried by cable.\\
        Question: What channels were removed from the network in March of 2007?\\
        Note: Your answer should be directly extracted from the passage and be a single entity, name, or number, not a sentence. If the passage doesn't contain a suitable answer, respond with 'unanswerable'.\\
        Answer: the basic channels\\
        
        Passage: Following the Cretaceous–Paleogene extinction event, the extinction of the dinosaurs and the wetter climate may have allowed the tropical rainforest to spread out across the continent. From 66–34 Mya, the rainforest extended as far south as 45°. Climate fluctuations during the last 34 million years have allowed savanna regions to expand into the tropics. During the Oligocene, for example, the rainforest spanned a relatively narrow band. It expanded again during the Middle Miocene, then retracted to a mostly inland formation at the last glacial maximum. However, the rainforest still managed to thrive during these glacial periods, allowing for the survival and evolution of a broad diversity of species.\\
        Question: Savannah areas expanded over the last how many years?\\
        Note: Your answer should be directly extracted from the passage and be a single entity, name, or number, not a sentence. If the passage doesn't contain a suitable answer, respond with 'unanswerable'.\\
        Answer: 34 million\\
        
        Passage: It is conjectured that a progressive decline in hormone levels with age is partially responsible for weakened immune responses in aging individuals. Conversely, some hormones are regulated by the immune system, notably thyroid hormone activity. The age-related decline in immune function is also related to decreasing vitamin D levels in the elderly. As people age, two things happen that negatively affect their vitamin D levels. First, they stay indoors more due to decreased activity levels. This means that they get less sun and therefore produce less cholecalciferol via UVB radiation. Second, as a person ages the skin becomes less adept at producing vitamin D.\\
        Question: As a person gets older, what does the skin produce less of?\\
        Note: Your answer should be directly extracted from the passage and be a single entity, name, or number, not a sentence. If the passage doesn't contain a suitable answer, respond with 'unanswerable'.\\
        Answer: vitamin D\\
        
        Passage: In 1066, Duke William II of Normandy conquered England killing King Harold II at the Battle of Hastings. The invading Normans and their descendants replaced the Anglo-Saxons as the ruling class of England. The nobility of England were part of a single Normans culture and many had lands on both sides of the channel. Early Norman kings of England, as Dukes of Normandy, owed homage to the King of France for their land on the continent. They considered England to be their most important holding (it brought with it the title of King—an important status symbol).\\Question: What battle took place in the 10th century?\\
        Note: Your answer should be directly extracted from the passage and be a single entity, name, or number, not a sentence. If the passage doesn't contain a suitable answer, respond with 'unanswerable'.\\
        Answer: unanswerable\\
        
        Passage: Dendritic cells (DC) are phagocytes in tissues that are in contact with the external environment; therefore, they are located mainly in the skin, nose, lungs, stomach, and intestines. They are named for their resemblance to neuronal dendrites, as both have many spine-like projections, but dendritic cells are in no way connected to the nervous system. Dendritic cells serve as a link between the bodily tissues and the innate and adaptive immune systems, as they present antigens to T cells, one of the key cell types of the adaptive immune system.\\
        Question: What is named for its resemblance to dendritic cells?\\
        Note: Your answer should be directly extracted from the passage and be a single entity, name, or number, not a sentence. If the passage doesn't contain a suitable answer, respond with 'unanswerable'.\\
        Answer: unanswerable\\
        
        Passage: \{CONTEXT\}\\
        Question: \{QUESTION\}\\
        Note: Your answer should be directly extracted from the passage and be a single entity, name, or number, not a sentence. If the passage doesn't contain a suitable answer, respond with 'unanswerable'.\\
        Answer:
    \end{mdframed}
    \caption{Five-shot English Prompt using SQuAD 2.0 examples.}
    \label{fig:appendix:five_shot_prompt_english}
\end{figure*}

\subsection{LLM Postprocessing}
\label{sec:appendix:llm-postprocessing}

In the zero-shot setting, Llama tends to generate more than just the answer span; for instance, "Based on the passage, the answer to the question is: [...] The passage states: [...]". This is not the only pattern in which the model phrases the answer. To capture as many as possible, we split the text at the semicolon and take the part that follows the semicolon. The answer and potential text passages to back it up are, in most cases, separated by newlines. We use this to remove text that is not part of the answer span: If there is no semicolon, we take the whole text output as the answer. These problems are particularly pronounced with Llama 2 and occur less with Llama 3.

\subsection{Isolated Domain and Language Transfer}
\label{sec:appendix:isolated-domain-and-language-transfer}
To further investigate the effect of domain and language, we investigate domain transfer and language transfer isolated. Overall, these configurations did not improve performance.

\paragraph{Isolated Language Transfer}
To explore the isolated language transfer, we eliminated domain variation. We provide GPT-3.5 with a prompt in a different language and examples in the language from the same domain. German gets a Turkish prompt, Turkish gets a Chinese prompt and Chinese gets a German prompt. This results in no improved performance as can be seen in Table \ref{table:appendix:results-ablation-language-transfer}.

\paragraph{Isolated Domain Transfer}
To perform only domain transfer, we eliminate language variation in the chat. We let native speakers translate the prompts to the target languages (German, Turkish, Chinese). Thus, in the zero-shot scenario, the model gets the system prompt, passage, question and note in the target language. 
For the five-shot evaluation, the examples come from the same language but from a different domain. The results are shown in Table \ref{table:appendix:results-ablation-domain-transfer}.

\begin{table}[]
    \small
    \center
    \begin{tabular}{llcc}
        \toprule
            & & \multicolumn{2}{c}{five-shot} \\
            & & F1 & EM \\
        \midrule
        \mr{3}{German}  & Creative Writing  & 52.89& 43.07\\
                        & Product Reviews   & 54.77& 45.27\\
                        & News              & 59.34& 49.33\\
        \cmidrule(lr){1-2}\cmidrule(lr){3-4}
        \mr{3}{Turkish}  & Creative Writing & 55.36& 46.33\\
                        & Product Reviews   & 58.96& 43.47\\
                        & News              & 55.72& 41.67\\
        \cmidrule(lr){1-2}\cmidrule(lr){3-4}
        \mr{3}{Chinese}  & Creative Writing & 37.80& 37.13\\
                        & Product Reviews   & 40.18& 39.67\\
                        & News              & 30.81& 26.73\\
        \midrule
        Average& & 49.54& 41.41\\
        \bottomrule
    \end{tabular}
    \caption{Five-shot language transfer with \texttt{gpt-3.5-turbo-0613}. The prompts contain examples from the target domain. The prompt for German is written in Chinese, for Turkish in German and for Chinese in Turkish.}
    \label{table:appendix:results-ablation-language-transfer}
    \vspace{-1.5em}
\end{table}

\begin{table}[]
    \small
    \center
    \resizebox{\linewidth}{!}{%
    \begin{tabular}{llcccc}
        \toprule
            & & \multicolumn{2}{c}{zero-shot} & \multicolumn{2}{c}{five-shot} \\
            & & F1 & EM & F1 & EM \\
        \midrule
        \mr{3}{German}  & Creative Writing & 55.49& 46.07& 58.11& 46.67\\
                        & Product Reviews & 54.88& 44.47& 52.01& 39.33\\
                        & News & 60.88& 49.87& 60.46& 48.40\\
        \cmidrule(lr){1-2}\cmidrule(lr){3-4}\cmidrule(lr){5-6}
        \mr{3}{Turkish} & Creative Writing & 36.93& 28.07& 52.71& 44.27\\
                        & Product Reviews & 45.52& 30.40& 51.88& 35.80\\
                        & News & 43.59& 28.93& 54.54& 39.13\\
        \cmidrule(lr){1-2}\cmidrule(lr){3-4}\cmidrule(lr){5-6}
        \mr{3}{Chinese} & Creative Writing & 44.38& 44.33& 47.98& 47.93\\
                        & Product Reviews & 41.65& 41.60& 41.45& 41.40\\
                        & News & 34.27& 32.47& 31.14& 30.20\\
        \midrule
        Average& & 46.40& 38.47& 50.03&41.4\\
        \bottomrule
    \end{tabular}
    }
    \caption{Domain transfer with \texttt{gpt-3.5-turbo-0613}: This table evaluates zero-shot and five-shot prompts written in the target language. The five-shot prompt for creative writing contains examples from M2QA news, the prompt for news from M2QA product reviews and the prompt for product reviews from M2QA creative writing.}
    \label{table:appendix:results-ablation-domain-transfer}
\end{table}

\begin{table}[]
    \centering
    \small
    \begin{tabularx}{\linewidth}{lYYY}
        \toprule
        Language & Annotators Kept & Annotators Rejected & Questions Checked  \\
        \midrule
        German & 66 & 10 & 760 \\ 
        Turkish & 32 & 12 & 440 \\ 
        Chinese & 33 & 9 & 420 \\ 
        \bottomrule
    \end{tabularx}
    \caption{Number of annotators we kept and how many we have rejected due to poor quality. For each annotator, we checked 10 questions. If at least two questions were of poor quality, i.e. did not follow our guidelines, the annotator got rejected. The last column shows how many of the accepted and rejected questions we checked in total for quality.}
    \label{table:quality-assurance-checked-questions}
    \vspace{-2.5mm}
\end{table}

\subsection{Investigating German GPT-3.5 Answers}
\label{sec:appendix-gpt-answer-investigation}
To gain further insights into GPT-3.5's performance, we chose to sample some hard questions and include a case study to analyze them. We manually inspected German questions for which all four GPT-3.5 setups achieved an F1 score lower than 25, which are 942 questions in total (20.9\% of all German QA instances). From these questions, we randomly sampled 50 questions to analyze the responses of all GPT-3.5 models we evaluated, i.e. the responses of the zero-shot and five-shot \texttt{gpt-3.5-turbo-0301} and \texttt{gpt-3.5-turbo-0613}. We found that in 72\% of the cases, the question and answer are correctly annotated in the data, but the model either makes erroneous predictions (58\%) or generates a correct answer instead of extracting it (14\%). We further identified issues with inconsistent annotations (22\%, i.e. 4.6\% of all German data), questions with multiple plausible answers (4\%), and the evaluation metric (2\%). We provide some representative answers in Table \ref{table:appendix:gpt-35-answers}. The full evaluation can be found in the M2QA GitHub repository.\footnote{\url{https://github.com/UKPLab/m2qa/tree/main/Experiments/LLM_evaluation}}

\begin{table*}[]
    \centering
    \tiny
    \begin{tabularx}{\textwidth}{lp{0.05\textwidth}p{0.55\textwidth}p{0.05\textwidth}p{0.08\textwidth}p{0.1\textwidth}}
        \toprule
        ID & Question & Passage Text & Expected Answer & Answer by five-shot gpt-3.5-turbo-0613 & Reason Why Answer Is Wrong\\
        \midrule
         \rotatebox[origin=l]{270}{de\_news\_125-0\_q0}& Welche Position spielt Marc Janko?& Vorsichtig gab sich auch Stürmer Marc Janko: Uns Spielern ist die Schwere des Gegners bewusst, wir hatten in Moldawien ein sehr hektisches und schwieriges Spiel, erinnerte er an den knappen 2:1-Auswärtssieg im Oktober. Es war eine Partie, in der die österreichische Nationalelf mit der Spielweise Moldaus so manches Problem hatte. Am Samstag wird der moldauische Teamchef Alexandru Curtianu auf zwei Schlüsselspieler verzichten müssen: Der 28-jährige Abwehrstratege Alexandru Epureanu – mit 58 Einsätzen einer der erfahrensten Teamspieler – fällt wegen eines Kreuzbandrisses monatelang aus. Der Kapitän, der für Medipol Basaksehir, den Zwölften der türkischen Süperlig, aufläuft und früher zur Stammformation von Dinamo Moskau gehörte, ist mit einem Marktwert von 4,5 Millionen Euro der wertvollste Spieler der moldauischen Nationalelf.& Stürmer& unanswerable&Question and expected answer fine; model made a wrong prediction\\
         \midrule
         
         \rotatebox[origin=l]{270}{de\_books\_2\_61\_q0} & Was schlägt Klamm vor? & „Wollen wir es nun trotzdem versuchen, dennoch versuchen, ein Bündnis zu schließen? Wollen Sie meine Frau werden? Können Sie dem Vorurteil begegnen, daß ich nicht als der Freiherr von Klamm auftrete, der als Mann einer sehr reichen Frau lediglich die Zeit stiehlt und im Müßiggang lebt, sondern ein Geschäft, ein Gewerbe betreibt, arbeitet, schafft, fördert, maßvoll lebt, den rechten Lebensgewinn in dem Verkehr mit gleichgesinnten, wertvollen Personen erblickt, die denselben Anschauungen huldigen, so überlegen Sie meinen abermaligen Antrag! Aber gönnen Sie mir auch — verzeihen Sie das viele — das Gelöbnis, daß Sie lediglich mein sein und bleiben wollen, daß Sie“ — Klamm sprach's mit einem sanften, gewinnenden Lächeln — „keine anderen Götter haben wollen, neben mir!“ & ein Bündnis zu schließen & Klamm suggests getting married.& Question and expected answer fine; model generated a correct answer instead of extracting it (often in english)\\
         \midrule
         
         \rotatebox[origin=l]{270}{de\_news\_142-0\_q2}& Wer ist bald fuer NGOs zustaendig?& -H Chinas Parteibürokratie sieht das anders. Ihr Verbot scheint Teil jüngster Willkür-Maßnahmen in der reideologisierten Innenpolitik Chinas zu sein, um die Zivilgesellschaft unter ihre Kontrolle zu bringen. Die Behörden statuierten mit der Schließung der Fraueninitiative, die auch von der Ford-Stiftung unterstützt wird, ein Exempel für alle zu eng mit dem Ausland verbundenen NGOs. Peking steht kurz vor Verabschiedung eines repressiven neuen Gesetzes für Auslands-NGOs. Betroffen sind Bürgerinitiativen, Stiftungen und Vereine. Sie sollen sich neu registrieren lassen und müssen ihre Arbeitspläne und Finanzen offenlegen. Künftig sollen sie der administrativen Kontrolle der Polizei unterstehen, statt wie bisher den Zivilämtern. & Sie sollen sich neu registrieren lassen & die Polizei&low-quality annotations\\
         \midrule
         
         \rotatebox[origin=l]{270}{de\_news\_26-1\_q0}& Welcher Partei gehört Heiko Maas an? & Bundesjustizminister Heiko Maas hat den Handgranaten-Anschlag scharf verurteilt. Das Ausmaß der Gewalt ist erschreckend, erklärte der SPD-Politiker am Freitag in Berlin. Die Täter dürfen nicht ungestraft davonkommen. Sie müssen konsequent ermittelt und bestraft werden, forderte er. Die Zunahme der Angriffe auf Flüchtlinge sei dramatisch. Sprengkörper auf Flüchtlingsheime fliegen heute schon, wir dürfen nicht abwarten, bis es die ersten Toten gibt. Ähnlich äußerte sich der Zentralrat Deutscher Sinti und Roma. Dieser feige Anschlag zeigt, dass gewaltbereite Rechtsextremisten durch ihre Taten den Frieden in unserer Gesellschaft gefährden und uns auseinanderdividieren wollen, erklärte sein Vorsitzender Romani Rose. Umso mehr gelte es, für die Demokratie und den Rechtsstaat einzustehen. Besonders Politiker trügen hierbei eine große Verantwortung. Die populistische Rhetorik in der Asyldebatte führt dazu, dass Ängste bei der Bevölkerung geschürt werden, kritisierte Rose. & SPD-Politiker & SPD &problem with the evaluation metric\\
         \midrule

         \rotatebox[origin=l]{270}{de\_review\_22\_q2}& Ist dieses Produkt emp\-feh\-lens\-wert? & Ich bin begeistert. Dieses kleine Ding ist die Lösung auf all meinen Reisen. Wie oft ich mich geärgert habe, dass die sch*** Adapter nicht passen und ich lauter Netzgeräte einstecken musste, damit ich Handy, Kamera usw laden kann. Die Lösung kann so einfach sein. Absolut empfehlenswert. Zusätzliches Plus: Das Gerät besitzt eine eigene Sicherung (was in so manchem Ländern durchaus sinnvoll ist) und eine Ersatzsicherung wird gleich mitgeliefert. Würde auch 6* geben wenn ich könnte.& Absolut & empfehlenswert&multiple answers would be correct\\
        
        \bottomrule
    \end{tabularx}
    \vspace{-3mm}
    \caption{Samples of questions that five-shot \texttt{gpt-3.5-turbo-0613} failed at, along with the reason.}
    \label{table:appendix:gpt-35-answers}
    \vspace{-4mm}
\end{table*}

\section{Results on SQuADv2 and XQuAD}
To show that our baselines and adapter-based setups do not only work on M2QA, we evaluated them also on SQuADv2 \cite{rajpurkar-etal-2018-know}, and XQuAD \cite{artetxe-etal-2020-cross}. Important to note is, that XQuAD only contains answerable questions. The results are presented in Table \ref{tab:results_squad_xquad_m2qa}.

\section{Annotation Process}
\label{sec:appendix:annotation_process_quality}
The number of annotators that were rejected vs. accepted during the annotation process and how many questions were checked in total is shown in Table \ref{table:quality-assurance-checked-questions}.

\section{Data Annotation Platform}
\label{sec:appendix-platform-ui}
To be able to fulfil all of our requirements, we have developed our own annotation platform. The source code, including the tutorial, i.e. the instructions for the crowdworkers, is published in the same GitHub repository as the dataset\footnote{\url{https://github.com/UKPLab/m2qa/tree/main/Website}}. We used GitHub Copilot\footnote{\url{https://github.com/features/copilot}} as AI assistance during coding. The crowdworkers first land on an overview page, then complete the tutorial and finally annotate data for M2QA. An annotation session consists of the tutorial and the annotation of 11 passages. If an annotator completed the tutorial in a previous session, it is optional, and they are assigned 12 passages. We assume that one annotation session results in a total of 1 hour of work. An evaluation after 65 annotation sessions showed that the crowdworkers took a median of 59 minutes. We pay crowdworkers £9 per annotation session, which is Prolifics recommended pay per hour.

The tutorial consists of 3 steps in which the annotator is subsequently introduced to the task and learns to use the data annotation platform:
\begin{itemize}
    \item[1.] On the first page, they get an introduction to the annotation task.
    \item[2.] Then they learn what makes good answerable questions and what to avoid when creating them.
    \item[3.] Last, they learn what requirements good unanswerable questions must fulfil and what to avoid when creating them.
\end{itemize}

\begin{figure*}[]
    \centering
    \includegraphics[width=\textwidth]{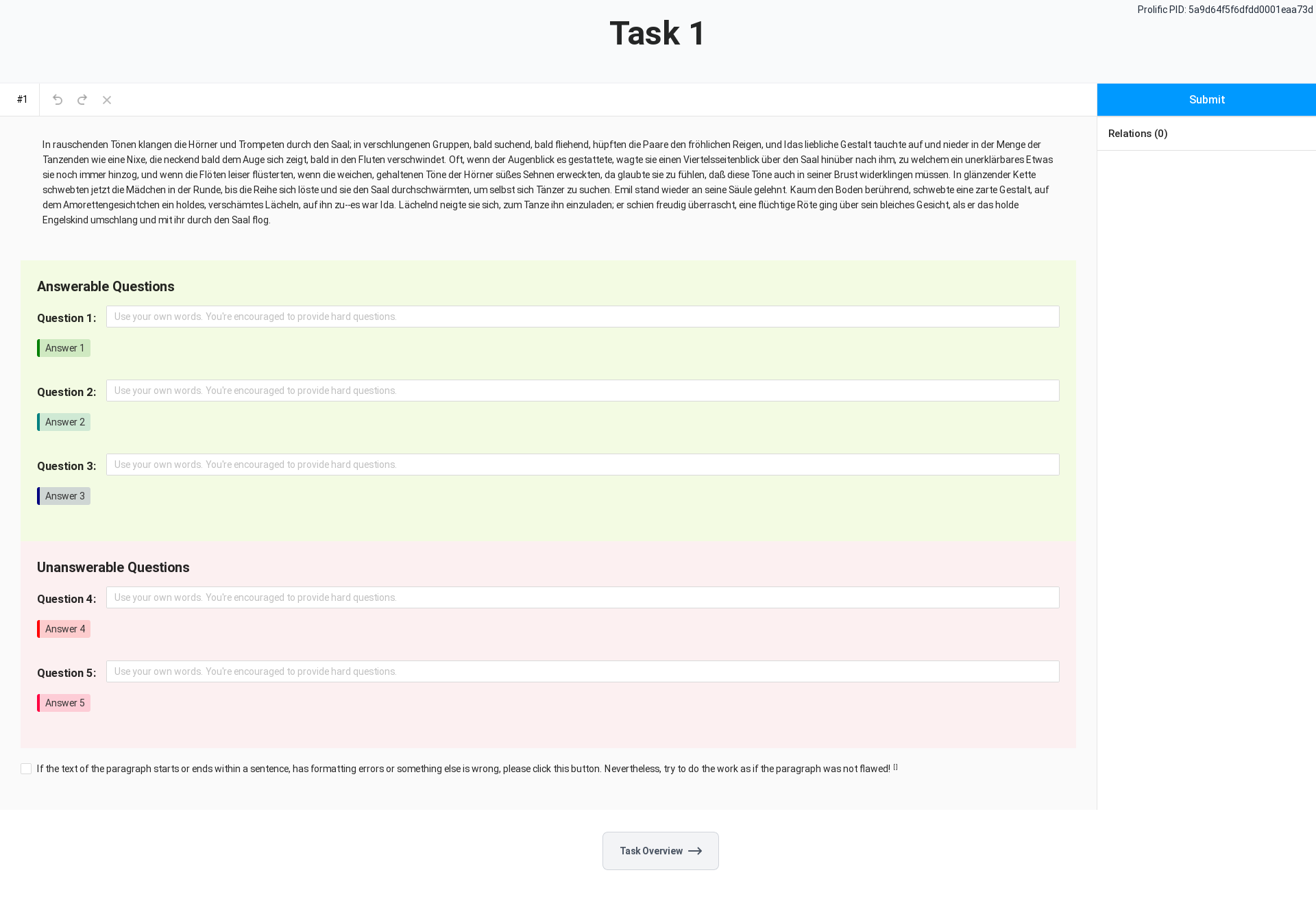}
    \caption{Screenshot of the interface that annotators use to write answerable and unanswerable questions and mark the respective answer span. Our interface is based on the Label Studio Frontend \cite{tkachenko_label_2020}.}
    \label{image:appendix-annotate}
\end{figure*}

Figure \ref{image:appendix-annotate} shows the interface that annotators use to annotate passages. Following SQuAD \cite{rajpurkar-etal-2016-squad}, we encourage annotators to pose hard questions in their own words. Since the wide adoption of LLM chatbots, the concern has arisen that crowdworkers could increasingly use LLMs to generate data instead of creating it themselves \cite{veselovsky2023artificial}. By disabling copy-pasting and requiring manual highlighting of the answer spans, we believe that using a ChatBot is not efficient in our setup. We found no evidence of the usage of LLMs during our quality checks.

\begin{table*}[]
    \centering
    \resizebox{\textwidth}{!}{%
    \begin{tabular}{llcccccccccccc}
        \toprule
                    &           & \multicolumn{2}{c}{SQuADv2} & \multicolumn{2}{c}{XQuAD English} & \multicolumn{2}{c}{XQuAD German} & \multicolumn{2}{c}{XQuAD Turkish} & \multicolumn{2}{c}{XQuAD Chinese} & \multicolumn{2}{c}{M2QA Total Average} \\
         Model      &           & F1    & EM    & F1    & EM    & F1    & EM    & F1    & EM & F1    & EM   & F1    & EM           \\
         \midrule
         \XLMRBase  &           (0-shot) & 74.72& 70.96& 71.86& 62.18& 53.62& 40.59& 48.73& 36.97& 42.30& 35.13& 37.73&31.59\\
         \XLMRDomain    &       (0-shot) & 73.44& 70.66& 67.27& 57.56& 37.26& 29.41& 16.93& 12.52& 18.78& 18.87& 36.36&33.26\\
         \cmidrule(lr){1-2}\cmidrule(lr){3-4} \cmidrule(lr){5-12} \cmidrule(lr){13-14}
         MAD-X+Domain &             (0-shot) & 75.50& 72.32& 69.34& 59.92& 51.24& 39.66& 46.57& 35.21& 41.72& 34.20& 41.42& 37.68\\
         MAD-X$^2$  &   (0-shot) & 77.03& 74.11& 68.70& 59.83& 51.65& 40.34& 21.24& 16.47& 30.16& 24.71& 41.89& 38.28\\ 
         \midrule
         Llama 2-chat (13b) &(0-shot) & 38.98& 30.02& 64.74& 46.47& 46.10& 30.67& 25.12& 14.62& 19.09& 6.47& 17.95&10.09\\
         Llama 3-instruct (8b) &(0-shot) & 57.05& 52.18& 77.86& 64.03& 66.93& 50.92& 57.76& 39.92& 53.69& 47.14& 44.54&29.59\\
         \cmidrule(lr){1-2}\cmidrule(lr){3-4} \cmidrule(lr){5-12} \cmidrule(lr){13-14}
         gpt-3.5-turbo-0301 &(0-shot)& 67.34& 59.56& 76.50& 60.00& 68.35& 47.65& 58.50& 35.13& 41.29& 35.71& 47.08&37.50\\
         gpt-3.5-turbo-0613 &(0-shot)& 72.92& 68.50& 78.20& 66.39& 68.16& 52.35& 61.65& 43.78& 60.95& 58.15& 53.11&45.00\\
         \cmidrule(lr){1-2}\cmidrule(lr){3-4} \cmidrule(lr){5-12} \cmidrule(lr){13-14}
         Aya-23 (8b) &(0-shot)& 77.49& 74.87& 78.94& 70.84& 69.04& 56.55& 63.73& 49.75& 62.08& 58.07& 51.61&44.16\\
         
         \bottomrule
    \end{tabular}
    }
    \vspace{-2mm}
    \caption{Results of the base models and adapter-based methods on SQuAD, XQuAD and M2QA.}
    \vspace{-3mm}
    \label{tab:results_squad_xquad_m2qa}
\end{table*}

\end{document}